\documentclass{article}

\PassOptionsToPackage{numbers, compress}{natbib}

\usepackage[preprint]{main}

\usepackage{iftex}
\ifPDFTeX
  \usepackage[utf8]{inputenc}
  \usepackage[T1]{fontenc}
\else
  \usepackage{fontspec}
\fi

\usepackage{hyperref}
\usepackage{url}
\usepackage{booktabs}
\usepackage{amsfonts}
\usepackage{nicefrac}
\usepackage{microtype}
\usepackage[table]{xcolor}
\usepackage{amsmath}
\usepackage{amssymb}
\usepackage{amsthm}
\usepackage{graphicx}
\usepackage{float}
\usepackage{algorithm}
\usepackage{algorithmic}
\usepackage{hyperref}
\usepackage{enumitem}
\usepackage{hyperref}
\usepackage{fontawesome}
\usepackage{twemojis}
\usepackage{algorithm}
\usepackage{setspace}

\ifPDFTeX\else
\fi

\usepackage{multirow}
\usepackage{multicol}
\usepackage{subcaption}
\usepackage{wrapfig}
\newtheorem{proposition}{Proposition}

\definecolor{bluelight}{HTML}{e8efff}
\definecolor{orgLight}{HTML}{fbdacf}
\definecolor{gainGreen}{HTML}{008000}

\definecolor{lightgreen}{RGB}{0,150,0}
\definecolor{myred}{RGB}{200,0,0}
\usepackage{mathtools}
\newcommand{\cimp}[2]{%
  \ensuremath{%
    #1%
    \mathrlap{_{\scriptstyle\textcolor{lightgreen}{\textbf{#2}}}}%
  }%
}
\usepackage{array}

\newcolumntype{A}{>{\centering\arraybackslash}p{1.15cm}}

\title{On the Position Bias of On-Policy Distillation}

\vspace{-10pt}
\author{
Yan Xie$^1$\thanks{Equal Contribution. \texttt{\{yanxie940, zsj200454\}@gmail.com}}
\quad Sijie Zhu$^{1}$\footnotemark[1]
\quad Tiansheng Wen$^{2}$ 
\quad Bo Chen$^1$\thanks{Corresponding Authors: Bo Chen (bchen@mail.xidian.edu.cn) and Yifei Wang (yifeiwg@amazon.com). This work was conducted outside of Amazon.}
\quad Yifei Wang$^{3}$\footnotemark[2] \\
$^1$Xidian University \quad
$^2$Georgia Institute of Technology \quad
$^3$Amazon AGI SF Lab \\  
}

\begin{document}

\maketitle
\vspace{-5pt}

\begin{center}
  \vspace{-23pt}
  \href{https://yannx1e.github.io/IW-OPD/}{\texttwemoji{1f4c4}\;Blog}
  \quad\quad
  \href{https://huggingface.co/IW-OPD}{\texttwemoji{1f917}\;Hugging Face}
  \quad\quad
  \href{https://github.com/YannX1e/Importance-Weighted-On-Policy-Distillation}{\faGithub\;Github}
  \end{center}
  \vspace{-2pt}

\begin{abstract}
  On-Policy Distillation (OPD) improves the learning efficiency of standard reinforcement learning through dense, token-level supervision from teachers. In the standard KL objective of OPD, token-level losses are uniformly averaged, implying equal weights for all tokens. However, we discover that not all tokens are created equal: as student rollouts grow longer, they deviate further from the teacher's distribution, leading to degraded supervision quality at later positions. As a result, OPD using only the first 30\% of tokens can perform comparably to using all tokens, whereas OPD using only the last 30\% of tokens barely learns anything. In this work, we provide a principled understanding of this issue through the lens of constrained optimization. Based on these insights, we derive Importance-Weighted On-Policy Distillation (IW-OPD), in which the weight assigned to each token depends on the accumulated discrepancy between the student's and teacher's distributions, naturally upweighting earlier tokens and downweighting later ones with larger deviations. We show that IW-OPD converges significantly faster than OPD, with better learning efficiency, and achieves better final performance than standard OPD in both same-size and cross-scale settings, improving performance by $6.9$ points on AIME-2025.
\vspace{-15pt}
\end{abstract}
\vspace{-3pt}
\begin{figure}[H]
  \centering
  \begin{subfigure}[t]{0.45\textwidth}
    \centering
    \includegraphics[width=\linewidth]{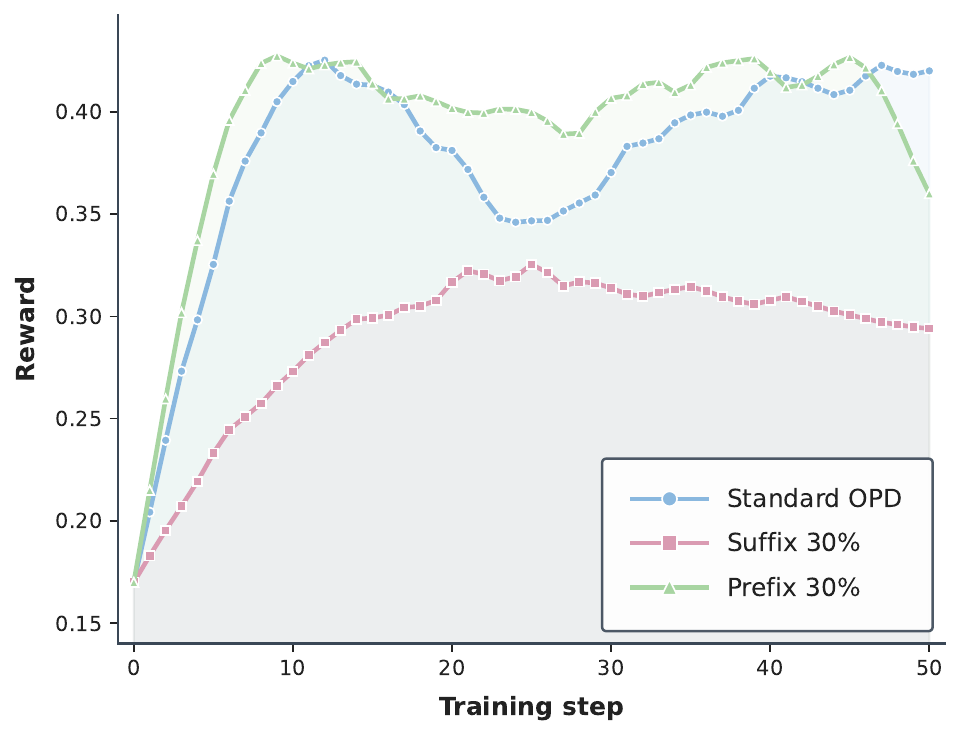}
    \caption{OPD training with the same token budget but supervision applied to different token positions.}
    \label{fig:context-corruption-a}
  \end{subfigure}
  \quad
  \begin{subfigure}[t]{0.42\textwidth}
    \centering
    \includegraphics[width=\linewidth]{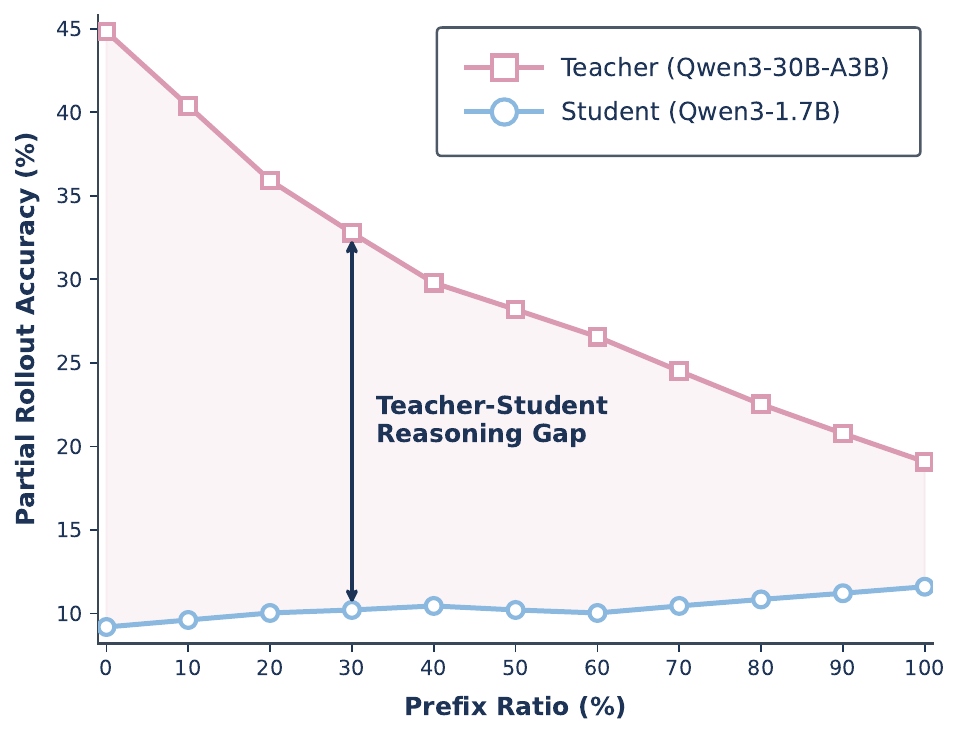}
    \caption{Teacher–student gap in final accuracy when conditioned on student-generated prefixes.}
    \label{fig:context-corruption-b}
  \end{subfigure}
  \caption{\textbf{Position Bias in OPD training.} \textbf{(a)} With the same 30\% token budget, training on the prefix part of each response matches or exceeds full token Standard OPD, whereas training on the suffix part fails to learn effectively. Student: Qwen3-0.6B, Teacher: Qwen3-4B-Instruct-2507. \textbf{(b)} Teacher and student accuracy are measured by the probability of reaching a correct answer from a given student-generated prefix. Student model maintains a low accuracy, while the teacher model’s mean@32 accuracy of eventually reaching the correct answer drops rapidly toward the student level as the student-generated prefix becomes longer.}
  \label{fig:context-corruption}
\vspace{-5pt}
\end{figure}
\vspace{-5pt}

\begin{figure}[h]
  \centering
  \begin{subfigure}[t]{0.4\textwidth}
    \centering
    \includegraphics[width=\linewidth]{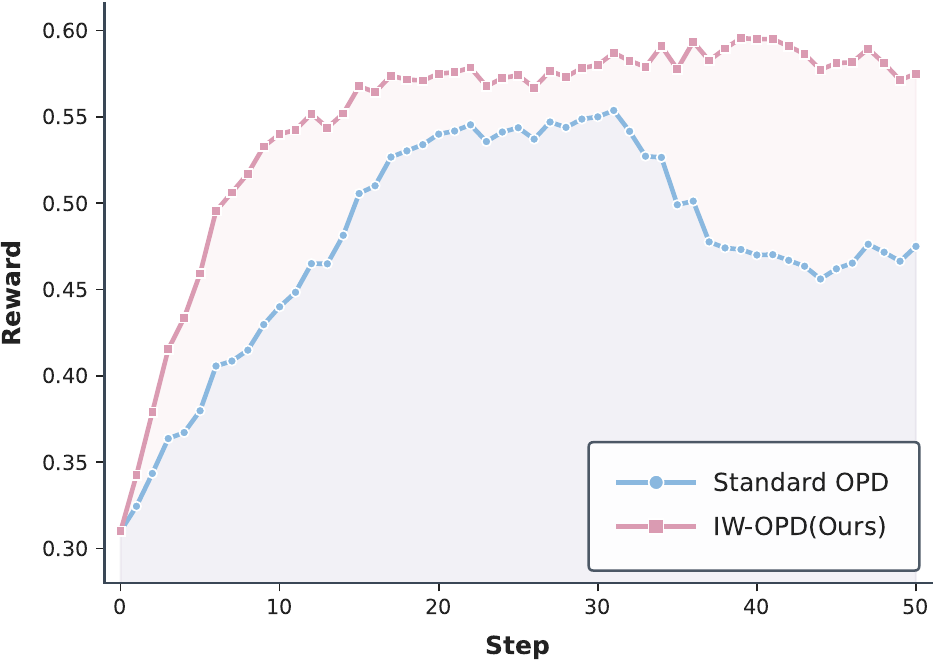}
    \caption{AIME25 accuracy during training. Teacher: Qwen3-4B, Student: Qwen3-1.7B.}
    \label{fig:compare-by-acc}
  \end{subfigure}
  \qquad
  \begin{subfigure}[t]{0.4\textwidth}
    \centering
    \includegraphics[width=\linewidth]{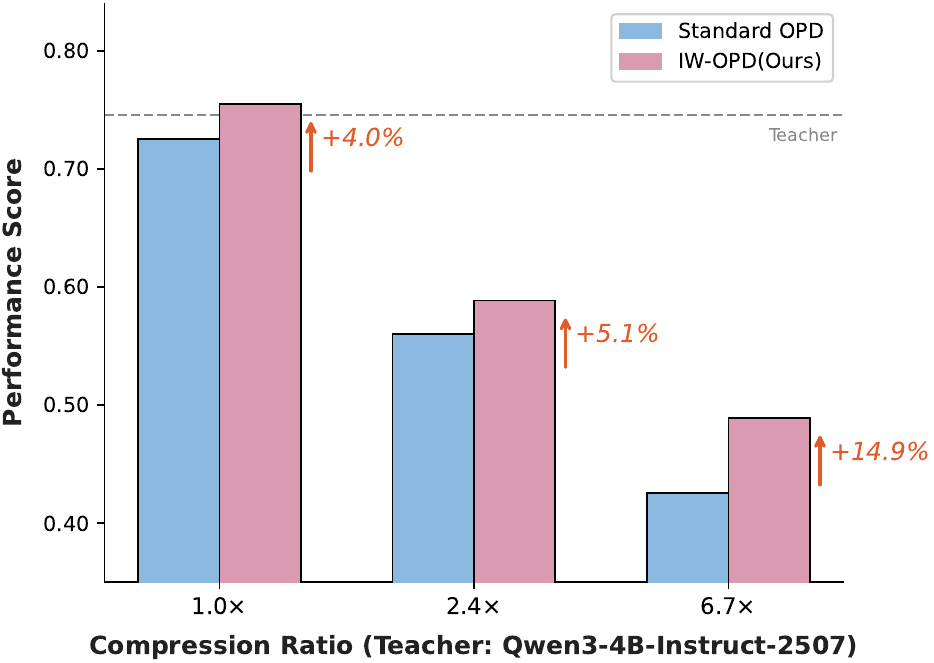}
    \caption{Final accuracy vs.\ compression ratio (teacher parameters / student parameters)}
    \label{fig:compare-by-bar}
  \end{subfigure}
  \caption{\textbf{IW-OPD improves both sample efficiency and final performance.} \textbf{(a)} AIME 2025 accuracy during training: IW-OPD converges faster and achieves better final performance than Standard OPD. \textbf{(b)} Final accuracy across student scales distilled from the same teacher; the IW-OPD advantage grows from $+4.0\%$ at $1.0\times$ compression to $+14.9\%$ at $6.7\times$.}
  \label{fig:context-corruption}
\vspace{-5pt}
\end{figure}

\section{Introduction}
\label{sec:intro}

On-Policy Distillation (OPD) trains a student on its own rollouts, while a stronger teacher provides dense token-level supervision at the prefixes visited by the student~\citep{agarwal2024gkd,gu2024minillm,lu2025,qwen3,xiao2026}, substantially improving learning efficiency over sparse trajectory-level rewards in LLM post-training~\citep{cui2025,rethinking_credit2026}.

An OPD objective uniformly aggregates per-token KL divergence as in standard knowledge distillation. However, it overlooks the nature of OPD, where samples are generated from a weak student, which often produces erroneous outputs that are out-of-distribution (OOD) for the teacher model. As shown in Figure~\ref{fig:context-corruption-b}, teacher can still provide reliable prediction when rolling out from  early student tokens, but its performance also deteriorates quickly at longer student rollouts, indicating that these prefixes have drifted away from the teacher distribution and it can provide limited value on it. This clear trend reveals a \textit{position bias} in OPD: early tokens in student rollouts should receive high-weights since it's high-quality, while later tokens should be down-weighted. 
A further controlled study confirms this intuition:  as shown in Figure~\ref{fig:context-corruption-a}, with the same 30\% token budget, OPD with only 30\% prefixes matches or exceeds full OPD, whereas OPD with 30\% suffix provides little benefit.

These observations suggest that OPD should be viewed as an allocation problem under a finite local-update budget. Since each update can move the student policy only a limited distance, the update should spend more gradient budget on prefixes where teacher supervision is still compatible with the student's trajectory. We formalize this intuition through a constrained local projection toward the teacher. Solving this constrained problem yields a closed-form optimal policy whose sample weights are governed by the teacher-to-student likelihood ratio. This ratio explains the observed \textit{Position Bias} phenomenon: once student rollouts move the trajectory away from the teacher-preferred reasoning path, the prefix ratio decreases, and optimal policy naturally reduces the sampling probability of downstream tokens.
IW-OPD (\textbf{I}mportance-\textbf{W}eighted \textbf{O}n-\textbf{P}olicy \textbf{D}istillation) implements this principle by reweighting token-level distillation terms with prefix-importance weights induced by the constrained projected objective. The method requires no additional teacher evaluations beyond standard OPD and reduces to standard OPD when the extra weighting is removed. Experiments show faster convergence and stronger final performance (Figure~\ref{fig:context-corruption}), with AIME25 gains over OPD reaching $+6.9$ points at step 10 and $+1.7$ points at convergence. Moreover, IW-OPD makes stronger teachers more sample-efficient and yields larger relative gains as students become smaller. Accordingly, this paper makes three contributions:

\begin{enumerate}[leftmargin=*, itemindent=0pt]
\item We identify the \textit{position bias} phenomenon in OPD and explain it from a constrained-optimization perspective. This view shows why teacher-compatible prefixes dominate useful supervision. (\S\ref{sec:token-uniformity}).

\item We propose IW-OPD as an efficient OPD objective with token-level importance estimated from the discrepancy between the teacher and the student models (\S\ref{sec:method}).

\item We demonstrate that in pratice, IW-OPD consistently improves OPD with faster convergence and stronger final performance, and that its advantage scales with teacher–student mismatch: stronger teachers become more sample-efficient, while smaller students obtain larger gains (\S\ref{sec:experiments}).

\end{enumerate}

\section{Preliminaries}
\label{sec:prelim}

Let $\mathcal{D}$ denote the prompt distribution, $\pi_\theta$ the student policy, and $\pi_T$ the teacher policy. For a prompt $x$ and response $y = (y_1,\dots,y_T)$, the trajectory-level distributions decompose autoregressively:
\begin{equation}
\pi_\theta(y | x) = \prod_{t=1}^T \pi_\theta(y_t | x, y_{<t}), \qquad \pi_T(y | x) = \prod_{t=1}^T \pi_T(y_t | x, y_{<t}).
\label{eq:autoregressive}
\end{equation}

\paragraph{On-Policy RL.} Standard RLVR (e.g., GRPO~\citep{shao2024deepseekmath}) samples trajectories from the current policy and optimizes a trajectory-level reward:
\begin{equation}
\mathcal{J}_{\text{RL}}(\theta) = \max_\theta \, \mathbb{E}_{x \sim \mathcal{D},\, y \sim \pi_\theta(\cdot | x)} \big[ r(x, y) \big],
\label{eq:rl}
\end{equation}
where $r(x,y)$ is obtained from a reward model~\citep{internlm2,rlhf-workflow,skywork-rm-v2} or a verifier~\citep{deepseekr1,code-r1,deepcritic,orz}. The policy gradient takes the form
\begin{equation}
\nabla_\theta \mathcal{J}_{\text{RL}} = \mathbb{E}_{x,\, y \sim \pi_\theta} \bigg[ \sum_{t=1}^{T} A_t \, \nabla_\theta \log \pi_\theta(y_t | x, y_{<t}) \bigg],
\label{eq:rl-grad}
\end{equation}
where the advantage $A_t$ computed from the trajectory reward assigns the same credit to every token.

\paragraph{On-Policy Distillation (OPD).} OPD~\citep{agarwal2024gkd,gu2024minillm,lu2025} replaces the sparse trajectory-level reward with dense token-level supervision from a teacher model $\pi_T$~\citep{qwen3,xiao2026}:
\begin{equation}
\mathcal{J}_{\mathrm{OPD}}(\theta)
= \max_\theta \, -D_{\mathrm{KL}}(\pi_\theta||\pi_T)=  -\mathbb{E}_{x,\, y \sim \pi_\theta}\sum_{t=1}^T\log \frac{\pi_\theta(y_t|x, y_{<t})}{\pi_T(y_t|x, y_{<t})},
\label{eq:opd-obj}
\end{equation}
where $y=[y_1,\dots,y_T]\sim \pi_\theta(y|x)$ denotes a sampled answer from the student $\pi_\theta$.
In practice, OPD decomposes \textit{sequence-level} objective and uses a \textit{token-local} semi-gradient that treats the sampled prefixes as fixed~\citep{xiao2026,lu2025} as in Eq.~\eqref{eq:rl-grad}:
\begin{equation}
\nabla_\theta \mathcal{J}_{\text{OPD}} \approx \mathbb{E}_{x,\, y \sim \pi_\theta} \bigg[ \sum_{t=1}^{T} 
A_t^{\text{OPD}}
\, \nabla_\theta \log \pi_\theta(y_t | x, y_{<t}) \bigg],
\label{eq:opd-grad}
\end{equation}
where OPD assigns a token-level advantage from the teacher--student distribution gap:
\begin{equation}
A_t^{\text{OPD}} \coloneqq -(\log \pi_\theta(y_t | x, y_{<t}) - \log \pi_T(y_t | x, y_{<t})).
\label{eq:opd-advantage}
\end{equation}
\section{Position Bias in On-Policy Distillation}
\label{sec:token-uniformity}

Standard OPD provides dense token-level supervision, but it aggregates all token-level KL terms uniformly in Eq.~\eqref{eq:opd-obj}. In this section, we observe that OPD actually has a clear \textit{position bias}: its early tokens are much more valuable for learning compared to its later tokens. We will show two interesting empirical phenomena in \S\ref{sec:position-bias} and provide a theoretical explanation in \S\ref{sec:understanding}.

\subsection{The Position Bias Phenomenon in OPD}
\label{sec:position-bias}

\paragraph{Early-token supervision drives OPD performance.} To evaluate the influence of token positions in OPD learning, we fix the supervision budget and vary only the supervised segment. Prefix-30 applies OPD to the first 30\% response tokens, while Suffix-30 applies OPD to the last 30\%; Standard OPD uses all valid tokens. All other training settings are unchanged (details in Appendix~\ref{app:experimental-details}).

Figure~\ref{fig:context-corruption-a} shows a strong asymmetry. Supervising only the prefix 30\% of tokens achieves performance comparable to standard OPD and consistently outperforms supervising only the suffix 30\%. In contrast, suffix-only supervision yields substantially lower rewards throughout training. These results indicate that OPD benefits primarily from early-token supervision, while supervision on later tokens alone provides limited gains.

\textbf{Teacher--student gap largely persists after OPD.}
Meanwhile, we also observe that even if the OPD objective tries to minimize the KL divergence between the teacher and student, the actual divergence between these two only decreases by 20\% even if training converges and the student performance saturates (shown in Fig.~\ref{fig:token-kl} and Fig.~\ref{fig:mean-kl}). It suggests that OPD training effectively only optimizes the student within a small local region. This could be because the OPD only optimizes on student-generated samples and this kind of RL training is known to produce minimal weight update\citep{wang2025forking,bigelow2025forkingpaths,lin2025criticaltokens,vassoyan2025ignore,ye2025disentangling}.

The two phenomena combined suggest an interesting learning landscape in OPD learning: \textit{it only optimizes the student distribution locally and most of the gains come from early prefix tokens}. Why does OPD have such a position bias and what does it imply for learning efficiency? We provide a theoretical explanation of this phenomenon in the next section.

\subsection{Understanding Position Bias from a Finite-Budget Allocation Perspective}
\label{sec:understanding}

As the empirical result in \S\ref{sec:position-bias} indicates that OPD only moves the student distribution within a small range, we can think of the actual OPD training as a constrained optimization problem, where the student distribution stays in a local region during training:
\begin{equation}
\min_{q}\ D_{\mathrm{KL}}({q}\|\pi_T)
\quad \mathrm{s.t.} \quad
D_{\mathrm{KL}}(q\|\pi_\theta) \le \rho,
\label{eq:constrained-proj}
\end{equation}
where $\pi_\theta$ denotes the student distribution, and $\rho$ denotes the effective local update budget measured by KL divergence. It can also be viewed as a trust-region objective where the policy is only updated within a trust region of radius $\rho$.
In fact, this constrained objective admits a \textbf{closed-form} solution $q^\star$, as revealed in the following proposition.
The proof can be found at Appendix~\ref{app:projection-solution}.

\begin{figure}[t]
  \centering
  \begin{subfigure}[t]{0.32\textwidth}
    \centering
    \includegraphics[width=\linewidth]{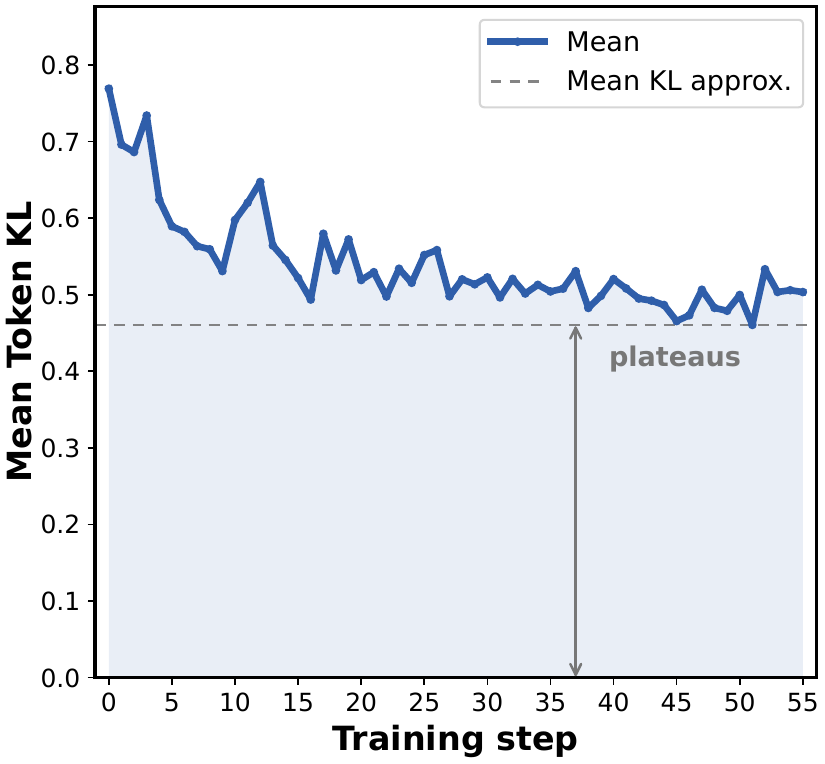}
    \caption{Mean token-level reverse KL across training steps.}
    \label{fig:token-kl}
  \end{subfigure}
  \hfill
  \begin{subfigure}[t]{0.32\textwidth}
    \centering
    \includegraphics[width=\linewidth]{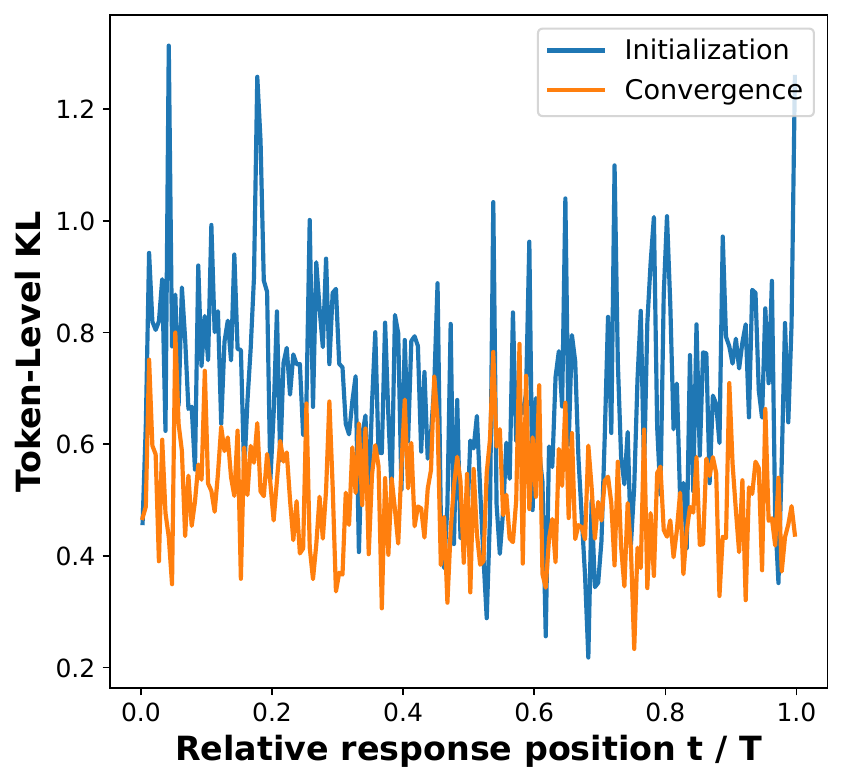}
    \caption{Token-level reverse KL before and after OPD training.}
    \label{fig:mean-kl}
  \end{subfigure}
  \hfill
  \begin{subfigure}[t]{0.32\textwidth}
    \centering
    \includegraphics[width=\linewidth]{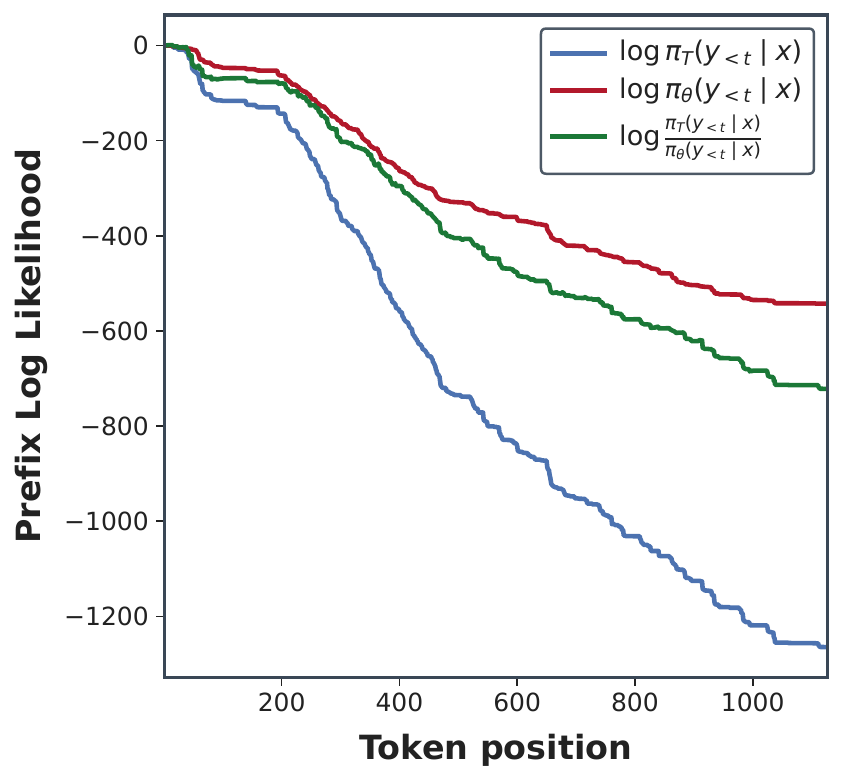}
    \caption{Log-likelihoods gap of student-sampled prefix vs. prefix length.}
    \label{fig:log-prefix}
  \end{subfigure}
  \caption{\textbf{Position Bias phenomena in OPD.} \textbf{(a)} The mean token-level KL decreases during OPD training but plateaus at a non-zero residual. \textbf{(b)}     Token-level reverse KL before and after OPD training. \textbf{(c)} Sequence-level log-probabilities of student-sampled prefixes under the student and teacher. Student: Qwen3-0.6B; Teacher: Qwen3-4B-Instruct.}
  \label{fig:token-selection}
  \vspace{-5pt}
\end{figure}

\begin{proposition}[Optimal Policy]
\label{prop:constrained-projection}
Given $\pi_{\theta}$ and $\pi_T$ with common support. In the local-update regime
$0<\rho<D_{\mathrm{KL}}(\pi_T\|\pi_\theta)$, the trust-region constraint in
Eq.~\eqref{eq:constrained-proj} is active and the unique solution is
\begin{equation}
q_\theta^\star(y)
=
\frac{\pi_\theta(y)}{Z_{\alpha}(\theta)}
\left(\frac{\pi_T(y)}{\pi_\theta(y)}\right)^{\alpha},
\label{eq:pro_1}
\end{equation}
which reweights the student policy $\pi_\theta$ by the likelihood ratio $r_\theta(y)=\pi_T(y)/\pi_\theta(y)$. Here, $Z_{\alpha}(\theta)
=
\mathbb{E}_{y\sim \pi_\theta}
\left[
r_\theta(y)^{\alpha}
\right]$ denotes the normalizing factor, and  $\alpha\in(0,1)$ is a constant depends on  $\rho$. 
\end{proposition}
\vspace{-3pt}
\textbf{The optimal policy $q^\star$ is proportional to likelihood ratio.}
This likelihood-based reweighting view in Proposition~\ref{prop:constrained-projection} explains why the position bias in Figure~\ref{fig:context-corruption-a} arises. 
Along student rollouts, the prefix probability under the student, \(\pi_\theta(y_{<t})\), usually decreases smoothly because the sequence is sampled from \(\pi_\theta\). 
In contrast, the same prefix probability under the teacher, \(\pi_T(y_{<t})\), can drop much faster once early student decisions move the reasoning path away from the teacher-preferred region, as illustrated in Figure~\ref{fig:log-prefix}. 
Therefore the prefix ratio $r_\theta(y_{<t})^\alpha$
tends to become smaller at later positions. 
In the constrained optimum, this decreasing ratio translates into smaller sampling weights for later, low-ratio prefixes and their associated tokens.
Equivalently, the constrained optimum naturally assigns position-dependent weights according to prefix compatibility with the teacher.
Thus, position bias reflects the intrinsic ratio-weighted structure of the constrained optimum.

\section{Importance-Weighted On-Policy Distillation}
\label{sec:method}

Based on the insights from \S\ref{sec:token-uniformity}, in this section, we propose a more efficient OPD objective, Importance-Weighted OPD, that leverages the position bias to learn more efficiently.

\subsection{Importance-Weighted OPD}
\label{sec:ideal-weight}
As discussed in \S\ref{sec:position-bias}, OPD can only optimize within a local region and the optimal policy it could attain is $q^\star_{\theta}$ (Eq.~\ref{eq:pro_1}), which reweights the base student policy $\pi_\theta$ with teacher--student gap measured by the likelihood ratio. As discussed in \S\ref{sec:understanding}, only tokens with relatively high likelihood ratios provide meaningful learning signals in OPD since the others have very low probability of being sampled. 

Motivated by this observation, we propose to directly optimize the divergence between the optimal policy $q^\star$ and the teacher, which directly emphasizes high-probability samples:
\begin{equation}
\mathcal{J}_{q_\theta^\star}=\max_\theta \; -D_{\mathrm{KL}}(q_{\theta}^\star \| \pi_T).
\label{eq:proj_obj}
\end{equation}
Since $q_{\theta}^\star$ is a reparameterized distribution induced by the student and teacher, we can further reparameterize this new learning objective by sampling from $\pi_\theta$ instead. More specifically, it will be equivalent to an Importance-Weighted OPD (IW-OPD) objective, as revealed in the next proposition.

\paragraph{Importance-weighted form of the projected objective.}
For clarity, we first fix the prompt $x$ and omit it from the notation.
We consider the non-trivial local-update regime
$0<\rho<D_{\mathrm{KL}}(\pi_T\|\pi_\theta)$ from
Proposition~\ref{prop:constrained-projection}. In this regime
$\alpha$ is induced by
the effective step-size budget $\rho$, and the trust-region constraint is
active as $D_{\mathrm{KL}}(q_{\theta}^{\star}\|\pi_\theta)=\rho$. Thus we have:
\begin{align}
D_{\mathrm{KL}}(q_{\theta}^{\star}\|\pi_T)
&=
\mathbb{E}_{q_{\theta}^{\star}}
\left[
\log \frac{q_{\theta}^{\star}(y)}{\pi_T(y)}
\right] \notag \\
&=
\mathbb{E}_{q_{\theta}^{\star}}
\left[
\log \frac{q_{\theta}^{\star}(y)}{\pi_\theta(y)}
\right]
+
\mathbb{E}_{q_{\theta}^{\star}}
\left[
\log \frac{\pi_\theta(y)}{\pi_T(y)}
\right] \notag\\
&=
\rho+
\mathbb{E}_{q_{\theta}^{\star}}
\left[
\log \frac{\pi_\theta(y)}{\pi_T(y)}
\right].
\label{eq:tilde_obj}
\end{align}

\begin{proposition}[Importance-Weighted OPD Objective]
\label{prop:iw-opd-objective}
By applying a change of measure from \(q_\theta^\star\) to \(\pi_\theta\) and substituting Eq.~\eqref{eq:pro_1}, we obtain an importance-weighted trajectory-level objective and define the following token-level surrogate (proof in Appendix~\ref{app:iw-opd-proof}):

\begin{equation}
\mathcal{J}_{\mathrm{IW}}^\star(\theta)
=
\max_\theta \;
-\mathbb{E}_{y\sim \pi_\theta}
\left[
\sum_{t=1}^{T}
\mathrm{sg} \left[ \tilde r_{t} \right]
\log
\frac{
\pi_\theta(y_t\mid y_{<t})
}{
\pi_T(y_t\mid y_{<t})
}
\right],
\quad
\tilde r_{t}
=
\frac{r_t}{Z_{\alpha,t}} .
\label{eq:pro_2}
\end{equation}
where \(r_t := r_\theta(y_{<t}) = \pi_T(y_{<t})/\pi_\theta(y_{<t})\) depends on $\alpha$ and denotes the prefix likelihood ratio at position \(t\), inherited from the trajectory-level ratio in Proposition~\ref{prop:constrained-projection}. The position-wise normalizer is $Z_{\alpha,t}
=
\mathbb{E}_{y_{<t}\sim\pi_\theta}
\left[
r_t
\right]$. $\mathrm{sg}[\cdot]$ is stop gradient operator.
\end{proposition}
Eq.~\eqref{eq:pro_2} shows that the Eq.~\eqref{eq:proj_obj} objective can be optimized using $\pi_\theta$-sampled rollouts as standard OPD, with each token-level KL term reweighted by a detached, normalized prefix-importance weight $\tilde r_{t}$. 
This weight, introduced by $q_\theta^\star$, becomes larger for teacher-compatible prefixes with high teacher--student likelihood ratio and smaller after accumulated teacher--student drift.

Similar to OPD, the gradient of IW-OPD can be written as a policy gradient form with importance-weighted advantage (proof in Appendix~\ref{app:standard-opd-semigradient}):
\begin{align}
\nabla_\theta \mathcal{J}_{\mathrm{IW}}^\star(\theta)
&\approx
\mathbb{E}_{y\sim \pi_\theta}
\left[
\sum_{t=1}^{T}
A_t^\text{IW-OPD} 
\nabla_\theta \log \pi_\theta(y_t\mid y_{<t})
\right],
\label{eq:iw-opd-gradient}
\end{align}
where
\begin{equation}
   A_t^\text{{IW-OPD}}
=
-
\mathrm{sg} \left[ \tilde r_{t} \right]
\left( \log \pi_\theta(y_t\mid y_{<t})
-
\log \pi_T(y_t\mid y_{<t}) \right). 
\end{equation}
With stop-gradient applied, the coefficient in Eq.~\eqref{eq:iw-opd-gradient} acts as a multiplicative weight on the OPD policy-gradient signal. 
Thus, the weighted gradient is the operational form of the constrained-optimization view in Eq.~\eqref{eq:proj_obj}: it corrects OPD's position bias by reallocating the finite update budget through  prefix-importance weights.

\begin{figure}[t]
  \centering
  \begin{subfigure}[t]{0.32\textwidth}
    \centering
    \includegraphics[width=\linewidth]{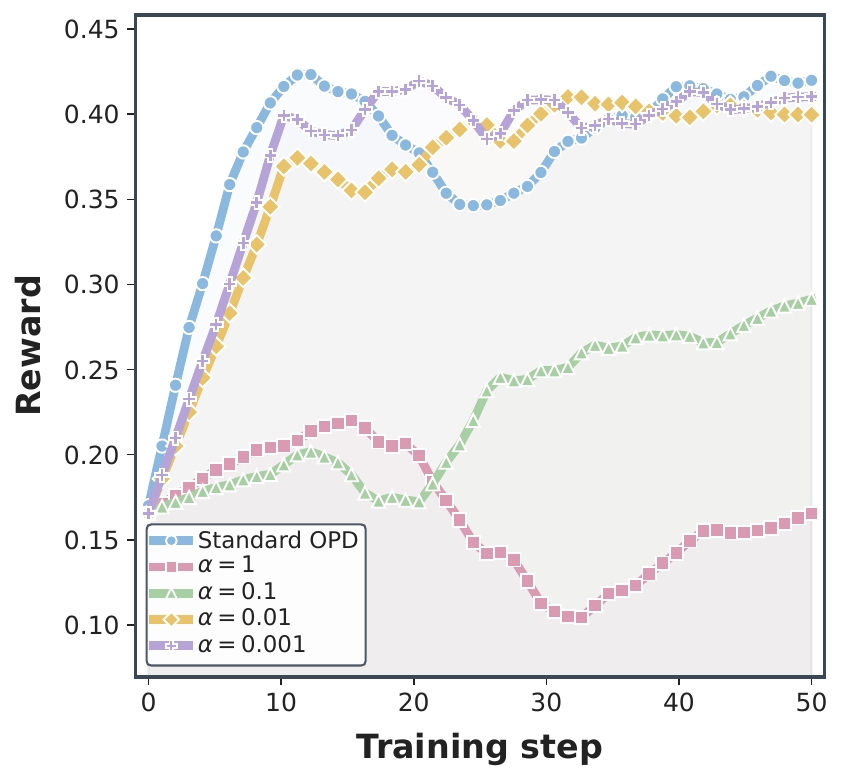}
    \caption{$\alpha$ in prefix-importance weights ablation.}
    \label{fig:ideal-weight-training}
  \end{subfigure}
  \hfill
  \begin{subfigure}[t]{0.32\textwidth}
    \centering
    \includegraphics[width=\linewidth]{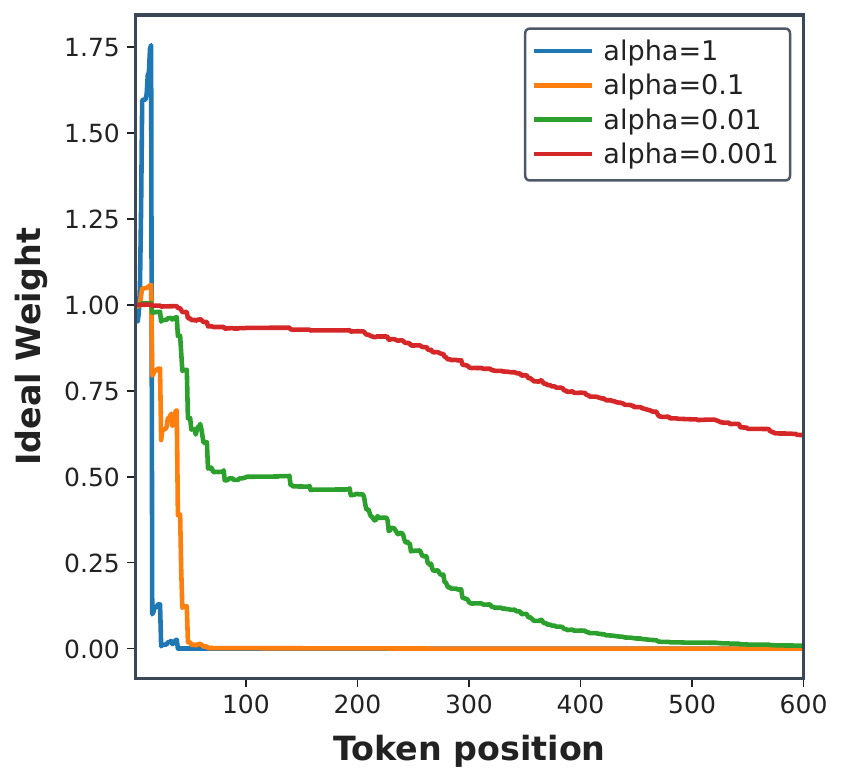}
    \caption{Prefix-importance weights visualization.}
    \label{fig:ideal-weight-visualization}
  \end{subfigure}
  \hfill
  \begin{subfigure}[t]{0.32\textwidth}
    \centering
    \includegraphics[width=\linewidth]{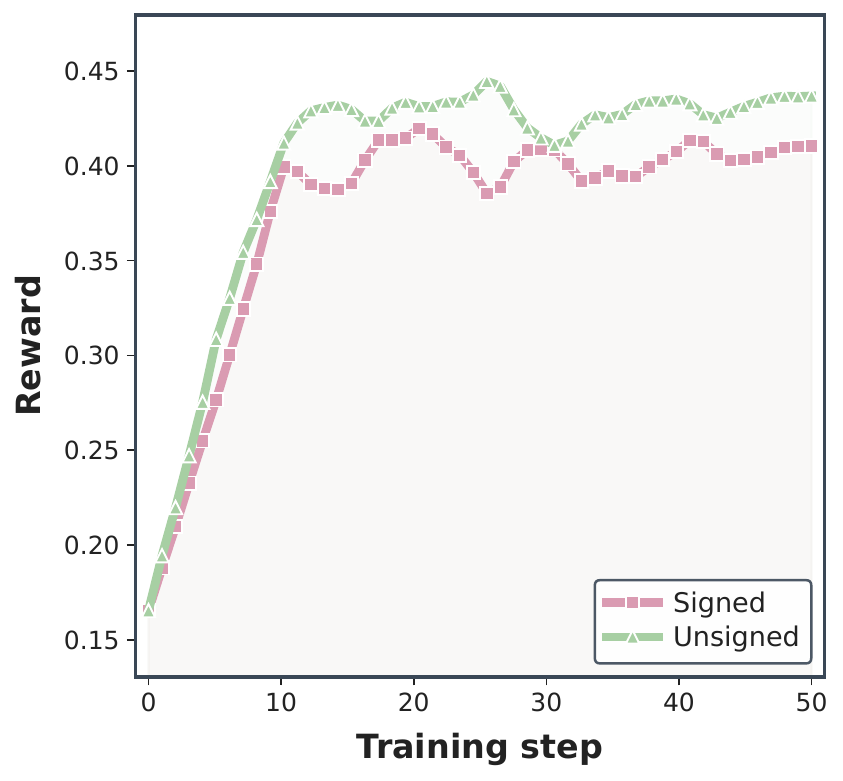}
    \caption{Signed vs. Unsigned prefix-importance weights ablation.}
    \label{fig:sign_vs_unsign}
  \end{subfigure}
  \caption{\textbf{From signed prefix ratio to unsigned prefix discrepancy.} \textbf{(a)} Directly using the ideal prefix ratio is sensitive to $\alpha$. \textbf{(b)} Token-wise visualization shows the desired overall downward trend, but also local rebounds caused by signed cancellation. \textbf{(c)} Replacing signed accumulation with the unsigned weight gives a more stable weighting signal. }
  \label{fig:main-fig-label}
  \vspace{-3pt}
\end{figure}

\subsection{Stable Token-level Importance Weight Estimate}
\label{sec:surrogate}

Eq.~\eqref{eq:iw-opd-gradient} provides a principled reweighting $\tilde{r}_{\alpha,t}\propto \pi_T(y_{<t})/\pi_\theta(y_{<t})$ for OPD. However, since the probability $\pi(y_{<t})=\Pi_{i=1}^{t-1}\pi(y_i|y_{<i})$ multiplies over longer sequence, the ratio will vary dramatically as $t$ grows, which introduces severe training instability. Below, we discuss several techniques helpful to stabilize this importance weight estimate (techniques are ablated in \S\ref{sec:ablation}).

\textbf{Small weight index $\alpha$.} A small $\alpha\to0$ can help flatten the difference, but it is still not sufficient. 
Fig.~\ref{fig:ideal-weight-training} shows strong sensitivity to $\alpha$: large values such as $\alpha=1$ and $\alpha=0.1$ substantially degrade training, while only very small values such as $\alpha=0.01$ and $\alpha=0.001$ roughly match standard OPD. 
Fig.~\ref{fig:ideal-weight-visualization} visualizes this numerical instability: $\alpha$ directly controls the scale and sharpness of the prefix weights, making the raw weights either overly concentrated or nearly flat. 

Beyond adjusting $\alpha$, we discover several effective strategies that stabilize the importance estimate.

\textbf{I. Stabilization via log scaling.} 
We first scale down the the probability gap by considering their ratio in the log space, which will preserve their order while mitigating the exponentially accumulated gaps:
\begin{equation}
\tilde{r}^\mathrm{log}_t=\log r_t = 
\alpha \sum_{k<t}\left(\log \pi_T(y_k|y_{<k})-\log \pi_\theta(y_k|y_{<k})\right) =
\alpha \sum_{k<t} A_k^{\mathrm{OPD}},
\label{eq:log-rt}
\end{equation}
which directly corresponds to the sum of token-level OPD advantages $A_k^{\mathrm{OPD}}$.

\textbf{II. Correcting positive-advantage tokens.}
Since the rollout tokens are sampled from the student, $\pi_\theta(y_k|y_{<k})$ is often relatively large on these tokens, and thus, as shown in Fig.~\ref{fig:log-prefix}, $\log \pi_T(y_k|y_{<k})-\log \pi_\theta(y_k|y_{<k})$ is mostly negative. However, for some tokens where the teacher assigns higher probability than the student, this term becomes positive. These terms can partially offset the accumulated negative prefix gap, 
as shown in Fig.~\ref{fig:ideal-weight-visualization}. In practice, we find it helpful to further reduce this cancellation effect by reverting the negative importance weights, leading to a sum of non-negative token-level discrepencies measured by $|A_k^{\mathrm{OPD}}|$:
\begin{equation}
\tilde{r}^\mathrm{abs}_t=\sum_{k<t}\left(\mathbb{I}[A_k^{\mathrm{OPD}}<0]A_k^{\mathrm{OPD}}
-\mathbb{I}[A_k^{\mathrm{OPD}}>0]A_k^{\mathrm{OPD}}\right)=
-\sum_{k<t}\left|A_k^{\mathrm{OPD}}\right|.
\end{equation}
We empirically compare both the original (Eq.~\ref{eq:log-rt}) and unsigned versions in Fig.~\ref{fig:sign_vs_unsign}, both variants are improve in practice, while the unsigned version yields higher p better results.

\textbf{III. Normalization.}
Although sign correction ensures monotonicity of importance weights as sequence grow, the  absolute scale could still vary significantly across samples. To mitigate this effect, we therefore apply a simple within-sample standardization. In particular, for a sample whose importance weights lie in $[d_T,0]$, where the maximum $0$ corresponds to the beginning of the sequence and the minimum $d_T$ corresponds to the end, we standardize the discrepancy to $[0,1]$ with
\begin{equation}
\tilde{r}^\mathrm{norm}_t=
\frac{d_t-d_{\text{min}}}{d_\text{max}-d_{\text{min}}}
\frac{-\sum_{k<t}|A_k^{\mathrm{OPD}}|-(-\sum_{k<T}|A_k^{\mathrm{OPD}}|)}{0-(-\sum_{k<T}|A_k^{\mathrm{OPD}}|)}=
1-
\frac{\sum_{k<t}\left|A_k^{\mathrm{OPD}}\right|}
{\sum_{k<T}\left|A_k^{\mathrm{OPD}}\right|}.
\end{equation}

\textbf{IV. Interpolation with OPD.} Finally, after normalization, end-of-sequence tokens will have low weights. We perform an interpolation with the original OPD to balance these two effects:
\begin{equation}
\tilde{r}^\mathrm{IW-OPD}_t=1+\gamma\cdot\tilde{r}^\mathrm{norm}_t=1+\gamma\left(1-
\frac{\sum_{k<t}\left|A_k^{\mathrm{OPD}}\right|}
{\sum_{k<T}\left|A_k^{\mathrm{OPD}}\right|}\right),
\end{equation}
where higher $\gamma\geq0$ indicates a higher contribution from the teacher-informed importance weights. In practice, we find $\gamma=0.5$ is a good default choice (Other parameters in Appendix~\ref{app:training-parameters}).

\begin{table}[t]
\centering
\caption{Evaluation results with \textbf{Qwen3-30B-A3B-Instruct-2507} as teacher (small student--teacher overlap). Math results are reported as mean@32 accuracy (\%). Methods with subscript ${10}$ are evaluated at training step 10.
\textbf{Bold} indicates the best result within each student group.}
\label{tab:main_results_30b}
\resizebox{\textwidth}{!}{
\begin{tabular}{ll|ccc|cc|A}
\toprule
\multirow{2}{*}{\textbf{Student}} & \multirow{2}{*}{\textbf{Method}} & \multicolumn{3}{c|}{\textbf{Math}} & \multicolumn{2}{c|}{\textbf{Code}} & \multirow{2}{*}{\textbf{Avg}} \\
& & AIME24 & AIME25 & HMMT25 & \phantom{0}HE+\phantom{0} & MBPP+ & \\
\midrule
\multicolumn{2}{l|}{\textcolor{gray}{\textit{Teacher Model}}} 
& \textcolor{gray}{74.7} 
& \textcolor{gray}{62.8} 
& \textcolor{gray}{44.2} 
& \phantom{0}\textcolor{gray}{86.6}\phantom{00}
& \textcolor{gray}{75.1} 
& \textcolor{gray}{68.7}  \\
\midrule
\multirow{5}{*}{Qwen3-4B}
& Base          
& 23.1
& 21.4  
& 10.0 
& \phantom{0}75.3\phantom{00}
& 64.5 
& 38.9 \\

& OPD$_{10}$    
& 51.2
& 42.4 
& 23.5 
& \phantom{0}76.2\phantom{00}
& 66.7 
& 52.0 \\

& \cellcolor{bluelight}IW-OPD$_{10}$ 
& \cellcolor{bluelight}\cimp{56.2}{+5.0}
& \cellcolor{bluelight}\cimp{49.3}{+6.9}
& \cellcolor{bluelight}\cimp{27.3}{+3.8}
& \cellcolor{bluelight}\phantom{0}\cimp{76.8}{+0.6}\phantom{00}
& \cellcolor{bluelight}\cimp{68.1}{+1.4}
& \cellcolor{bluelight}\cimp{55.5}{+3.5} \\

& OPD           
& 55.3
& 48.0 
& 27.1 
& \phantom{0}77.2\phantom{00}
& 69.1 
& 55.3 \\

& \cellcolor{orgLight}IW-OPD   
& \cellcolor{orgLight}\cimp{\textbf{57.5}}{+2.2}
& \cellcolor{orgLight}\cimp{\textbf{49.7}}{+1.7}
& \cellcolor{orgLight}\cimp{\textbf{28.7}}{+1.6}
& \cellcolor{orgLight}\phantom{0}\cimp{\textbf{78.7}}{+1.5}\phantom{00}
& \cellcolor{orgLight}\cimp{\textbf{70.9}}{+1.8}
& \cellcolor{orgLight}\cimp{\textbf{57.1}}{+1.8} \\
\midrule
\multirow{5}{*}{Qwen3-1.7B}
& Base          
& 13.4
& 11.0 
& 6.8 
& \phantom{0}59.6\phantom{00}
& 52.5 
& 28.7 \\

& OPD$_{10}$    
& 30.5
& 20.2 
& 14.4 
& \phantom{0}53.0\phantom{00}
& 52.6 
& 34.1 \\

& \cellcolor{bluelight}IW-OPD$_{10}$
& \cellcolor{bluelight}\cimp{33.0}{+2.5}
& \cellcolor{bluelight}\cimp{23.2}{+3.0}
& \cellcolor{bluelight}\cimp{15.6}{+1.2}
& \cellcolor{bluelight}\phantom{0}\cimp{55.5}{+2.5}\phantom{00}
& \cellcolor{bluelight}\cimp{54.8}{+2.2}
& \cellcolor{bluelight}\cimp{36.4}{+2.3} \\

& OPD           
& 34.6
& 28.7  
& 15.5 
& \phantom{0}64.6\phantom{00}
& 53.7 
& 39.4 \\

& \cellcolor{orgLight}IW-OPD        
& \cellcolor{orgLight}\cimp{\textbf{35.5}}{+0.9}
& \cellcolor{orgLight}\cimp{\textbf{29.5}}{+0.8}
& \cellcolor{orgLight}\cimp{\textbf{16.4}}{+0.9}
& \cellcolor{orgLight}\phantom{0}\cimp{\textbf{65.2}}{+0.6}\phantom{00}
& \cellcolor{orgLight}\cimp{\textbf{55.0}}{+1.3}
& \cellcolor{orgLight}\cimp{\textbf{40.3}}{+0.9} \\
\midrule
\multirow{5}{*}{Qwen3-0.6B}
& Base          
& 1.5
& 3.4 
& 1.3
& \phantom{0}28.2\phantom{00}
& 28.4 
& 12.6 \\

& OPD$_{10}$    
& 6.2
& 14.1 
& 6.5 
& \phantom{0}26.9\phantom{00}
& 23.1 
& 15.4 \\

& \cellcolor{bluelight}IW-OPD$_{10}$ 
& \cellcolor{bluelight}\cimp{7.8}{+1.6}
& \cellcolor{bluelight}\cimp{15.8}{+1.7}
& \cellcolor{bluelight}\cimp{7.6}{+1.1}
& \cellcolor{bluelight}\phantom{0}\cimp{28.4}{+1.5}\phantom{00}
& \cellcolor{bluelight}\cimp{27.5}{+4.4}
& \cellcolor{bluelight}\cimp{17.4}{+2.0} \\

& OPD           
& 11.0
& 17.8  
& 7.1 
& \phantom{0}29.6\phantom{00}
& 28.7 
& 18.8 \\

& \cellcolor{orgLight}IW-OPD        
& \cellcolor{orgLight}\cimp{\textbf{11.5}}{+0.5}
& \cellcolor{orgLight}\cimp{\textbf{19.3}}{+1.5}
& \cellcolor{orgLight}\cimp{\textbf{8.0}}{+0.9}
& \cellcolor{orgLight}\phantom{0}\cimp{\textbf{32.5}}{+2.9}\phantom{00}
& \cellcolor{orgLight}\cimp{\textbf{31.9}}{+3.2}
& \cellcolor{orgLight}\cimp{\textbf{20.2}}{+1.4} \\
\bottomrule
\end{tabular}
}
\vspace{-5pt}
\end{table}

\begin{table}[t]
\centering
\caption{Evaluation results with \textbf{Qwen3-4B-Instruct-2507} as teacher (large student--teacher overlap). Math results are reported as mean@32 accuracy (\%).
Methods with subscript ${10}$ are evaluated at training step 10.
\textbf{Bold} indicates the best result within each student group.}
\label{tab:main_results_4b}
\resizebox{\textwidth}{!}{
\begin{tabular}{ll|ccc|cc|A}
\toprule
\multirow{2}{*}{\textbf{Student}} & \multirow{2}{*}{\textbf{Method}} & \multicolumn{3}{c|}{\textbf{Math}} & \multicolumn{2}{c|}{\textbf{Code}} & \multirow{2}{*}{\textbf{Avg}} \\
& & AIME24 & AIME25 & HMMT25 & \phantom{0}HE+\phantom{0} & MBPP+ & \\
\midrule
\multicolumn{2}{l|}{\textcolor{gray}{\textit{Teacher Model}}} &
\textcolor{gray}{60.4} &
\textcolor{gray}{46.7} & \textcolor{gray}{31.0} &
\phantom{0}\textcolor{gray}{82.5}\phantom{00} & \textcolor{gray}{71.3} &
\textcolor{gray}{58.4} \\
\midrule
\multirow{5}{*}{Qwen3-4B}
& Base       
& 23.1
& 21.4 & 10.0 
& \phantom{0}75.3\phantom{00} 
& 64.5 & 38.9 \\
& OPD$_{10}$  
& 56.3
& 45.7 & 23.6 
& \phantom{0}76.0\phantom{00} 
& 66.1 & 54.0 \\
& \cellcolor{bluelight}IW-OPD$_{10}$
& \cellcolor{bluelight}\cimp{58.7}{+2.4}
& \cellcolor{bluelight}\cimp{46.7}{+1.0} 
& \cellcolor{bluelight}\cimp{25.0}{+1.4} 
& \cellcolor{bluelight}\phantom{0}\cimp{77.8}{+1.8}\phantom{00}
& \cellcolor{bluelight}\cimp{67.5}{+1.4} 
& \cellcolor{bluelight}\cimp{55.1}{+1.2} \\
& OPD 
& 56.5
& 46.3 & 24.4 
& \phantom{0}76.3\phantom{00} 
& 67.8 & 54.3 \\
& \cellcolor{orgLight}IW-OPD  
& \cellcolor{orgLight}\cimp{\textbf{58.7}}{+2.2}
& \cellcolor{orgLight}\cimp{\textbf{46.7}}{+0.4} 
& \cellcolor{orgLight}\cimp{\textbf{25.0}}{+0.6} 
& \cellcolor{orgLight}\phantom{0}\cimp{\textbf{77.9}}{+1.6}\phantom{00} 
& \cellcolor{orgLight}\cimp{\textbf{68.2}}{+0.4} 
& \cellcolor{orgLight}\cimp{\textbf{55.3}}{+1.0} \\
\midrule
\multirow{5}{*}{Qwen3-1.7B}
& Base     
& 13.4
& 11.0 & 6.8 
& \phantom{0}59.6\phantom{00} 
& 52.5 & 28.7 \\
& OPD$_{10}$    
& 33.4
& 24.7 & 11.3 
& \phantom{0}61.1\phantom{00} 
& 53.4 & 36.8 \\
& \cellcolor{bluelight}IW-OPD$_{10}$  
& \cellcolor{bluelight}\cimp{35.2}{+1.8}
& \cellcolor{bluelight}\cimp{25.9}{+1.2} 
& \cellcolor{bluelight}\cimp{13.2}{+1.9} 
& \cellcolor{bluelight}\phantom{0}\cimp{62.0}{+0.9}\phantom{00} 
& \cellcolor{bluelight}\cimp{54.0}{+0.6} 
& \cellcolor{bluelight}\cimp{38.1}{+1.3} \\
& OPD 
& 34.0
& 26.4 & 13.7 
& \phantom{0}61.5\phantom{00} 
& 53.7 & 37.9 \\
& \cellcolor{orgLight}IW-OPD   
& \cellcolor{orgLight}\cimp{\textbf{35.2}}{+1.2}
& \cellcolor{orgLight}\cimp{\textbf{27.1}}{+0.7} 
& \cellcolor{orgLight}\cimp{\textbf{15.3}}{+1.6} 
& \cellcolor{orgLight}\phantom{0}\cimp{\textbf{62.8}}{+1.3}\phantom{00} 
& \cellcolor{orgLight}\cimp{\textbf{54.9}}{+1.2} 
& \cellcolor{orgLight}\cimp{\textbf{39.1}}{+1.1} \\
\midrule
\multirow{5}{*}{Qwen3-0.6B}
& Base          
& 1.5
& 3.4 & 1.3 
& \phantom{0}28.2\phantom{00} 
& 28.4 & 12.6 \\

& OPD$_{10}$
& 11.1
& 17.1 
& 6.9 
& \phantom{0}26.8\phantom{00} 
& 31.0 
& 18.6 \\

& \cellcolor{bluelight}IW-OPD$_{10}$
& \cellcolor{bluelight}\cimp{12.4}{+1.3}
& \cellcolor{bluelight}\cimp{\textbf{19.0}}{+1.9} 
& \cellcolor{bluelight}\cimp{9.4}{+2.5} 
& \cellcolor{bluelight}\phantom{0}\cimp{28.1}{+1.3}\phantom{00} 
& \cellcolor{bluelight}\cimp{33.9}{+2.9} 
& \cellcolor{bluelight}\cimp{20.6}{+2.0} \\

& OPD      
& 11.8
& 17.1 
& 7.7 
& \phantom{0}29.8\phantom{00} 
& 33.3 
& 20.0 \\

& \cellcolor{orgLight}IW-OPD   
& \cellcolor{orgLight}\cimp{\textbf{13.6}}{+1.8}
& \cellcolor{orgLight}\cimp{\textbf{19.0}}{+1.9} 
& \cellcolor{orgLight}\cimp{\textbf{9.6}}{+0.9} 
& \cellcolor{orgLight}\phantom{0}\cimp{\textbf{31.6}}{+1.8}\phantom{00} 
& \cellcolor{orgLight}\cimp{\textbf{35.7}}{+2.4} 
& \cellcolor{orgLight}\cimp{\textbf{21.9}}{+1.9} \\
\bottomrule
\end{tabular}
}
\vspace{-5pt}
\end{table}

\begin{table}[t]
\centering
\caption{Evaluation results with \textbf{Qwen3-235B-A22B-Instruct-2507} as teacher. Math results are reported as mean@32 accuracy (\%).
\textbf{Bold} indicates the best result within each student group.}
\label{tab:scaling_results}
\resizebox{\textwidth}{!}{
\begin{tabular}{ll|ccc|cc|A}
\toprule
\multirow{2}{*}{\textbf{Student}} & \multirow{2}{*}{\textbf{Method}} & \multicolumn{3}{c|}{\textbf{Math}} & \multicolumn{2}{c|}{\textbf{Code}} & \multirow{2}{*}{\textbf{Avg}} \\
& & AIME24 & AIME25 & HMMT25 & \phantom{0}HE+\phantom{0} & MBPP+ & \\
\midrule
\multicolumn{2}{l|}{\textcolor{gray}{\textit{Teacher Model}}} 
& \textcolor{gray}{80.7}
& \textcolor{gray}{69.2}
& \textcolor{gray}{55.6}
& \textcolor{gray}{90.2}
& \textcolor{gray}{77.6}
& \textcolor{gray}{74.7} \\
\midrule
\multirow{3}{*}{Qwen3-30B-A3B}
& Base          
& 28.4
& 23.4
& 15.2
& 77.8
& 69.5
& 42.9 \\

& OPD 
& 69.5
& 56.7
& 38.4
& 82.1
& 71.3
& 63.6 \\

& \cellcolor{orgLight}IW-OPD  
& \cellcolor{orgLight}\cimp{\textbf{70.8}}{+1.3}
& \cellcolor{orgLight}\cimp{\textbf{58.9}}{+2.2}
& \cellcolor{orgLight}\cimp{\textbf{40.5}}{+2.1}
& \cellcolor{orgLight}\cimp{\textbf{83.5}}{+1.4}
& \cellcolor{orgLight}\cimp{\textbf{73.7}}{+2.4}
& \cellcolor{orgLight}\phantom{0}\cimp{\textbf{ 65.5}}{+1.9}\phantom{00} \\
\bottomrule
\end{tabular}
}
\vspace{-3pt}
\end{table}

\section{Experiments}

\label{sec:experiments}

\subsection{Setup}

We evaluate IW-OPD in two teacher regimes and three student scales. The students are Qwen3-4B, Qwen3-1.7B, and Qwen3-0.6B. The first teacher is Qwen3-4B-Instruct-2507, which gives a larger overlap setting within the same model family. The second teacher is Qwen3-30B-A3B-Instruct-2507, which gives a smaller overlap setting and tests whether prefix weighting remains useful when the student often leaves the teacher's preferred trajectory region.

Training uses DeepMath with difficulty at least 6, about 57K problems, for math, and Eurus-RL-Code, about 25K problems, for code. We evaluate math on AIME 2024, AIME 2025, and HMMT 2025. We evaluate code on HumanEval+ and MBPP+. The baselines are the student base model before distillation and standard OPD. All reported numbers are averaged over three random seeds.

\subsection{Main Results}
\paragraph{IW-OPD consistently improves OPD.} Tab.~\ref{tab:main_results_30b}, \ref{tab:main_results_4b} report the main evaluation results. 
IW-OPD consistently improves over standard OPD across all evaluated teacher--student pairs and benchmarks. 
In the two full evaluation regimes with Qwen3-4B and Qwen3-30B-A3B teachers, IW-OPD improves the final reported average for every student scale. 
The additional experiment using Qwen3-235B-A22B-Instruct-2507 as the teacher and Qwen3-30B-A3B as the student in Tab.~\ref{tab:scaling_results} further shows that the prefix-importance weighting remains effective in a much larger distillation setting, improving the math average by 1.9 points over OPD. Beyond vanilla OPD, Appendix~\ref{app:exopd} shows that IW-OPD can also be combined with other OPD variant that redesign the reward term, such as ExOPD~\citep{yang2026gopd}.

\paragraph{IW-OPD improves sample efficiency.}
The early checkpoints show that reweighting changes the efficiency of the update, not only the final endpoint. 
IW-OPD is ahead of OPD at step 10 in every reported teacher--student regime. 
The effect is especially clear in the Qwen3-30B-A3B $\rightarrow$ Qwen3-4B setting, where IW-OPD$_{10}$ improves the average score from 52.0 to 55.5 and AIME25 from 42.4 to 49.3. 
Notably, this step-10 checkpoint already matches the final OPD checkpoint in average performance. 
This supports the allocation view: by reducing the relative influence of drifted prefixes, IW-OPD spends more of each update on prefixes where teacher supervision can still redirect the student.

\begin{table}[t]
  \caption{\textbf{Ablation results on AIME25.} We isolate trajectory-adaptive prefix selection, unsigned discrepancy, and the effect of using the practical surrogate with a standard OPD blend.}
  \label{tab:ablation}
  \centering
  \begin{tabular}{lcc}
    \toprule
    Variant & AIME25 & $\Delta$ vs.\ OPD baseline \\
    \midrule
    \multicolumn{3}{l}{\textit{Prefix selection strategy}} \\
    Standard OPD & 43.3 & 0.0 \\
    Amplify fixed prefix & 43.7 & +0.4 \\
    Linear decay & 44.1 & +0.7 \\
    Manual curriculum & 44.8 & +1.5 \\
    \textbf{Cumulative-share (ours)} & 48.9 & +5.6 \\
    \midrule
    \multicolumn{3}{l}{\textit{Signed vs.\ unsigned discrepancy}} \\
    Signed $\sum A_k$ & 45.9 & +2.6 \\
    \textbf{Unsigned $\sum |A_k|$ (ours)} & 48.9 & +5.6 \\
    \midrule
    \multicolumn{3}{l}{\textit{Weight source and OPD blend}} \\
    Ideal weight only & 42.1 & -1.2 \\
    Ideal weight w. OPD blend & 43.9 & +0.6 \\
    Surrogate weight only & 46.2 & +2.9 \\
    \textbf{OPD blend (ours)} & 48.9 & +5.6 \\
    \bottomrule
  \end{tabular}
\vspace{-5pt}
\end{table}

\paragraph{IW-OPD improves efficiency as teachers scale up.}
We next isolate the effect of teacher scaling by fixing the student. For the Qwen3-4B student, using the stronger Qwen3-30B-A3B teacher gives a better final OPD average than using the Qwen3-4B teacher, 55.3 versus 54.3. However, standard OPD is less sample-efficient with the stronger teacher in the early stage: at step 10, distilled from 30B-A3B teacher reaches only 52.0 average score, lower than 54.0 from 4B teacher. This suggests that a stronger teacher can provide a better final target but also induces larger teacher--student trajectory mismatch, so uniform OPD needs more updates before the student can effectively benefit from its supervision. IW-OPD alleviates this inefficiency. In the same Qwen3-4B student setting, IW-OPD$_{10}$ with 30B-A3B teacher reaches 55.5 average score, surpassing IW-OPD$_{10}$ with 4B teacher at 55.1. On AIME25, the reversal is even clearer: standard OPD$_{10}$ with 30B-A3B teacher is behind the 4B teacher, 42.4 versus 45.7, whereas IW-OPD$_{10}$ makes the 30B-A3B teacher outperform the 4B teacher, 49.3 versus 46.7. These results show that IW-OPD makes stronger teachers more sample-efficient by reallocating training signal toward teacher-compatible prefixes.

\paragraph{IW-OPD benefits more as students scale down.}
We first isolate the effect of student scaling by fixing the teacher. With Qwen3-4B-Instruct-2507 as the teacher, the final average improvement of IW-OPD over OPD increases as the student becomes smaller: from +1.0 points for the 4B student, to +1.2 points for the 1.7B student, and to +1.9 points for the 0.6B student. In relative terms, these correspond to approximately +1.8\%, +3.2\%, and +9.5\%, respectively. The same trend is also visible at step 10, where the relative gains grow from +2.0\% to +3.5\% and then to +10.8\% as the student size decreases. These results indicate that IW-OPD is especially helpful in cross-scale distillation: when the student is much smaller than the teacher, student rollouts are more likely to drift away from the teacher-compatible region, and uniform OPD wastes more update budget on low-quality downstream supervision.

\paragraph{IW-OPD scales to stronger teachers and larger students.}
The experiment with Qwen3-235B-A22B-Instruct-2507 as the teacher in Table~\ref{tab:scaling_results} provides a large-scale stress test beyond the small-student regimes. 
Although standard OPD already gives a strong 30B student, IW-OPD still improves all reported math benchmarks, with gains of 1.3 points on AIME24, 2.2 points on AIME25, and 2.1 points on HMMT25. 
This indicates that prefix-importance weighting is not merely a remedy for weak students; it remains useful when distilling a very strong teacher into a capable student.

\subsection{Ablations}
\label{sec:ablation}

Table~\ref{tab:ablation} isolates three design choices: how token weights are assigned, how prefix discrepancy is measured, and whether the weighted term should be blended with standard OPD. We use Qwen3-0.6B as the student because this setting makes prefix selection most visible.

\paragraph{Adaptive prefix selection matters more than a fixed shape.}

Amplifying a fixed ratio (30\%) prefix gives only $+0.4$, and a hand-designed position schedule gives $+1.5$. After unsigned correction, $\tilde{r}^\mathrm{IW-OPD}$ becomes a monotonically decreasing weight. Accordingly, we test a direct linear-decay variant, which gives only $+0.7$. The cumulative-share rule gives $+5.6$. This ordering separates the benefit of early token preference from the benefit of trajectory adaptivity. A smooth preference for earlier positions is not enough. The useful boundary between reliable and unreliable prefixes changes across rollouts, so the weight must follow each trajectory's own discrepancy trace. Easy rollouts can keep high weights for longer, while hard rollouts should reduce downstream weights earlier.

\paragraph{Unsigned discrepancy is the better prefix compatibility proxy.}

Using signed accumulation gives $+2.6$, while the unsigned version gives $+5.6$. Signed terms can cancel even when the prefix has passed through several model disagreements. This cancellation makes a drifted prefix appear compatible with the teacher. The absolute statistic treats each disagreement as evidence that the student has moved away from the shared prefix region.

\paragraph{The surrogate should allocate extra budget rather than replace OPD.}

The ideal likelihood-ratio weight is useful as a derivation target but not as a literal training rule. Using it alone collapses performance, and blending it with OPD gives only a small gain. This matches the instability observed in Figure~\ref{fig:main-fig-label}: the raw ratio has high variance and is sensitive to signed cancellations along long trajectories. The practical surrogate is more robust because it preserves the desired ordering of prefixes while removing the unstable product scale. However, using the surrogate alone is still weaker than blending it with OPD. The blend keeps the standard dense OPD signal as a floor and allocates additional update budget to compatible prefixes, which is the intended role of IW-OPD.

\section{Related Work}
\label{sec:related}
\textbf{On-policy and token-selective distillation.} Classical KD~\citep{hinton2015distilling} trains on teacher-generated data and can suffer from exposure bias~\citep{bengio2015scheduled,arora2022exposure,ross2011dagger}. OPD supervises student-sampled rollouts: GKD~\citep{agarwal2024gkd} unifies on/off-policy data through $f$-divergences, MiniLLM~\citep{gu2024minillm} optimizes reverse KL with policy-gradient estimators, and recent OPD work studies on-policy teacher supervision and its extensions~\citep{li2026opd,opd_survey2026,yang2026gopd,zhao2026opsd}. Selective distillation further asks where supervision should be applied, using sequence-level curricula~\citep{wen2023fdistill,pocl2025,paced2026} or token-level weights based on frequency, difficulty, teacher confidence, and student learning state~\citep{gu2020tokenadaptive,liang2021tokenwise,huang2025selectkd,xie2025adakd,kim2026tsdkd}. IW-OPD follows this token-selective view and uses on-policy prefix compatibility as the weighting signal.

\textbf{Credit assignment and reweighted policy updates.} RLVR pipelines~\citep{deepseekr1,shao2024deepseekmath} must assign sparse outcome rewards over long reasoning traces. Process-supervision and process-reward methods provide denser step-level feedback~\citep{lightman2023prm,wang2024mathshepherd,cui2025}, and recent token-level analyses identify high-entropy forking tokens, critical tokens, and reasoning rather than boilerplate tokens as disproportionate drivers of learning~\citep{wang2025forking,bigelow2025forkingpaths,lin2025criticaltokens,vassoyan2025ignore,ye2025disentangling,rethinking_credit2026}. IW-OPD allocates dense teacher supervision by prefix compatibility. Its constrained-projection view connects to trust-region and proximal policy updates~\citep{kakade2001natural,schulman2015trpo,schulman2017ppo}, sequence-level importance correction in GSPO and online DPO~\citep{zheng2025gspo,rafailov2023dpo,xpo2025}, and geometric interpolation/R\'enyi midpoints~\citep{vanerven2014renyi,gmpo2026}.

\section{Discussion and Conclusion}
\label{sec:conclusion}

The key insight of this work is that OPD supervision is not uniformly reliable along a student-generated trajectory. This creates a \textit{position bias}: teacher supervision is often more useful near the beginning of the rollout than near the end. Such bias is not unique to OPD. It may also appear in many methods involving two autoregressive sequence models, such as on-policy RL where samples are drawn from the current policy but updates move toward a new policy. The issue is especially visible in OPD since the teacher--student distribution gap is not known in advance, so we cannot predefine where their trajectories remain compatible.

In conclusion, we identify position bias as a key inefficiency in On-Policy Distillation and explain it through a finite-budget local projection view. Motivated by this analysis, IW-OPD reallocates additional gradient budget toward teacher-compatible prefixes using a stable cumulative prefix-discrepancy weight, while keeping standard OPD as the dense supervision floor. Experiments across same-family, cross-scale, and stronger-teacher settings show that this simple modification improves both sample efficiency and final performance. More broadly, our results suggest that effective on-policy supervision should account not only for token-level disagreement, but also for the trajectory context (i.e., the prefix) during which that disagreement occurs.

\newpage

\bibliographystyle{plainnat}
\bibliography{references}

\newpage
\appendix

\section{Proofs and Derivations}
\label{app:proofs}

The derivations below fix a prompt $x$ unless otherwise stated and omit the
conditioning on $x$ for notational simplicity. We write $\pi_T$ for the teacher
next-token policy and also for its autoregressive trajectory distribution, as
in the main text. The local projected distribution is denoted by
$q_\theta^\star$, and the corresponding causal prefix weight is
$r_\theta$. All distributions are understood to be supported on the common
support where the relevant KL divergences are finite.

\subsection{Solution of the Constrained Projection}
\label{app:projection-solution}

We prove Proposition~\ref{prop:constrained-projection}. The constrained
projection is
\begin{equation}
q_{\theta}^\star
=
\arg\min_{q}
D_{\mathrm{KL}}(q\|\pi_T)
\quad
\mathrm{s.t.}
\quad
D_{\mathrm{KL}}(q\|\pi_\theta)\le \rho,
\end{equation}
together with the normalization constraint $\sum_y q(y)=1$.

If the trust-region constraint were inactive, the solution would be the
unconstrained minimizer $q=\pi_T$. This is infeasible when
$0<\rho<D_{\mathrm{KL}}(\pi_T\|\pi_\theta)$, so the constraint must be active in
the local-update regime.
Since $D_{\mathrm{KL}}(q\|\pi_T)$ is strictly convex in
$q$ on the common
support and the KL ball is convex, the KKT conditions identify the unique
optimum.

Introduce the Lagrangian
\begin{equation}
\mathcal{L}(q,\lambda,\mu)
=
D_{\mathrm{KL}}(q\|\pi_T)
+
\lambda
\left(
D_{\mathrm{KL}}(q\|\pi_\theta)-\rho
\right)
+
\mu
\left(
\sum_y q(y)-1
\right),
\end{equation}
where $\lambda\ge 0$ is the multiplier for the trust-region constraint and
$\mu$ is the multiplier for normalization. Expanding the KL terms gives
\begin{equation}
\mathcal{L}(q,\lambda,\mu)
=
\sum_y q(y)\log\frac{q(y)}{\pi_T(y)}
+
\lambda
\sum_y q(y)\log\frac{q(y)}{\pi_\theta(y)}
-
\lambda\rho
+
\mu
\left(
\sum_y q(y)-1
\right).
\end{equation}

Taking the functional derivative with respect to $q(y)$ and setting it to
zero yields
\begin{align}
0
&=
\frac{\partial \mathcal{L}}{\partial q(y)} \notag\\
&=
\log q(y)-\log \pi_T(y)+1
+
\lambda
\left(
\log q(y)-\log \pi_\theta(y)+1
\right)
+
\mu .
\end{align}
Rearranging terms,
\begin{equation}
(1+\lambda)\log q(y)
=
\log \pi_T(y)
+
\lambda \log \pi_\theta(y)
+
\mathrm{const},
\end{equation}
where the constant absorbs $1+\lambda+\mu$ and is independent of $y$.
Therefore,
\begin{equation}
\log q(y)
=
\frac{1}{1+\lambda}\log \pi_T(y)
+
\frac{\lambda}{1+\lambda}\log \pi_\theta(y)
+
\mathrm{const}.
\end{equation}
Define
\begin{equation}
\alpha
\coloneqq
\frac{1}{1+\lambda}
\in (0,1),
\qquad
1-\alpha
=
\frac{\lambda}{1+\lambda}.
\end{equation}
Exponentiating and normalizing gives
\begin{equation}
q_{\alpha}(y)
=
\frac{
\pi_\theta(y)^{1-\alpha}\pi_T(y)^\alpha
}{
Z_\alpha(\theta)
},
\qquad
Z_\alpha(\theta)
=
\sum_y
\pi_\theta(y)^{1-\alpha}\pi_T(y)^\alpha .
\end{equation}
Equivalently, with
\begin{equation}
r_\theta(y)=\frac{\pi_T(y)}{\pi_\theta(y)},
\qquad
Z_\alpha(\theta)
=
\mathbb{E}_{y\sim \pi_\theta}
\left[
r_\theta(y)^\alpha
\right],
\end{equation}
we can write
\begin{equation}
q_{\alpha}(y)
=
\frac{
\pi_\theta(y)r_\theta(y)^\alpha
}{
Z_\alpha(\theta)
}.
\end{equation}

It remains to identify the value of $\alpha$ induced by the radius $\rho$.
Let
\begin{equation}
\psi(\alpha)\coloneqq \log Z_\alpha(\theta).
\end{equation}
Then
\begin{equation}
\psi'(\alpha)
=
\mathbb{E}_{q_{\alpha}}
\left[
\log r_\theta(y)
\right].
\end{equation}
Therefore,
\begin{align}
D_{\mathrm{KL}}(q_{\alpha}\|\pi_\theta)
&=
\mathbb{E}_{q_{\alpha}}
\left[
\log\frac{q_{\alpha}(y)}{\pi_\theta(y)}
\right] \notag\\
&=
\mathbb{E}_{q_{\alpha}}
\left[
\alpha \log r_\theta(y)-\psi(\alpha)
\right] \notag\\
&=
\alpha\psi'(\alpha)-\psi(\alpha).
\end{align}
Since the constraint is active, $\alpha$ is determined by
\begin{equation}
\rho
=
D_{\mathrm{KL}}(q_{\alpha}\|\pi_\theta)
=
\alpha
\frac{\partial}{\partial \alpha}\log Z_\alpha(\theta)
-
\log Z_{\alpha}(\theta),
\label{eq:rho-alpha-relation}
\end{equation}
This is the implicit relation for $\alpha$. Also,
\begin{align}
\frac{d}{d\alpha}
D_{\mathrm{KL}}(q_{\alpha}\|\pi_\theta)
&=
\alpha\psi''(\alpha) \notag\\
&=
\alpha\,
\mathrm{Var}_{q_{\alpha}}
\left[
\log r_\theta(y)
\right]
\ge 0 .
\end{align}
Thus increasing $\rho$ increases the corresponding interpolation coefficient
$\alpha$ whenever $\pi_T\ne \pi_\theta$. Finally,
$D_{\mathrm{KL}}(q_0\|\pi_\theta)=0$ and
$q_1=\pi_T$, so
$D_{\mathrm{KL}}(q_1\|\pi_\theta)=D_{\mathrm{KL}}(\pi_T\|\pi_\theta)$. Hence
each $0<\rho<D_{\mathrm{KL}}(\pi_T\|\pi_\theta)$ induces
$\alpha\in(0,1)$ and
\begin{equation}
q_{\theta}^\star(y)
=
q_{\alpha}(y)
=
\frac{
\pi_\theta(y)r_\theta(y)^{\alpha}
}{
Z_{\alpha}(\theta)
}.
\end{equation}
This is Eq.~\eqref{eq:pro_1}.

When the budget is large enough that the teacher itself is feasible, the
constraint can be inactive, $\lambda=0$, and $\alpha=1$, which recovers
$q_{\theta}^\star=\pi_T$. \hfill$\square$

\subsection{Derivation of the Importance-Weighted OPD Objective}
\label{app:iw-opd-proof}

We prove Proposition~\ref{prop:iw-opd-objective}. Fix a prompt $x$ and let
$h_t=(x,y_{<t})$. We keep the conditioning on $x$ explicit in this appendix.
We work in the non-trivial local-update regime of
Proposition~\ref{prop:constrained-projection}, where
$0<\rho<D_{\mathrm{KL}}(\pi_T\|\pi_\theta)$ and the trust-region constraint is
active:
\begin{equation}
D_{\mathrm{KL}}
\left(
q_{\theta}^\star
\middle\|
\pi_\theta
\right)
=
\rho .
\end{equation}

The projected objective in Eq.~\eqref{eq:proj_obj} is
\begin{equation}
\mathcal{J}_{q_\theta^\star}(\theta;x)
=
\max_{q_\theta^\star}
-D_{\mathrm{KL}}
\left(
q_{\theta}^\star
\middle\|
\pi_T
\right).
\end{equation}
By adding and subtracting $\log \pi_\theta(y\mid x)$ inside the KL, we obtain
\begin{align}
D_{\mathrm{KL}}
\left(
q_{\theta}^\star
\middle\|
\pi_T
\right)
&=
\mathbb{E}_{y\sim q_{\theta}^\star(\cdot\mid x)}
\left[
\log
\frac{
q_{\theta}^\star(y\mid x)
}{
\pi_T(y\mid x)
}
\right] \notag\\
&=
\mathbb{E}_{y\sim q_{\theta}^\star(\cdot\mid x)}
\left[
\log
\frac{
q_{\theta}^\star(y\mid x)
}{
\pi_\theta(y\mid x)
}
\right]
+
\mathbb{E}_{y\sim q_{\theta}^\star(\cdot\mid x)}
\left[
\log
\frac{
\pi_\theta(y\mid x)
}{
\pi_T(y\mid x)
}
\right] \notag\\
&=
\rho
+
\mathbb{E}_{y\sim q_{\theta}^\star(\cdot\mid x)}
\left[
\log
\frac{
\pi_\theta(y\mid x)
}{
\pi_T(y\mid x)
}
\right].
\label{eq:app-projected-objective}
\end{align}
Thus, under a fixed local-update budget $\rho$, minimizing the projected KL is
equivalent up to the constant $\rho$ to minimizing the second term in
Eq.~\eqref{eq:app-projected-objective}.

Following Eq.~\eqref{eq:pro_1}, write the trajectory-level likelihood ratio as
\begin{equation}
r_\theta(y\mid x)
=
\frac{
\pi_T(y\mid x)
}{
\pi_\theta(y\mid x)
},
\qquad
Z_\alpha(\theta,x)
=
\mathbb{E}_{y\sim\pi_\theta(\cdot\mid x)}
\left[
r_\theta(y\mid x)^\alpha
\right].
\end{equation}
Then the optimal projected policy can be written as
\begin{equation}
q_{\theta}^\star(y\mid x)
=
\frac{
\pi_\theta(y\mid x) r_\theta(y\mid x)^\alpha
}{
Z_\alpha(\theta,x)
},
\qquad
\alpha\in(0,1).
\end{equation}
Therefore,
\begin{equation}
\frac{
q_{\theta}^\star(y\mid x)
}{
\pi_\theta(y\mid x)
}
=
\frac{
r_\theta(y\mid x)^\alpha
}{
Z_\alpha(\theta,x)
}.
\end{equation}
For any measurable function $f$, this gives the change-of-measure identity
\begin{equation}
\mathbb{E}_{y\sim q_{\theta}^\star(\cdot\mid x)}
[f(y)]
=
\frac{
\mathbb{E}_{y\sim\pi_\theta(\cdot\mid x)}
\left[
r_\theta(y\mid x)^\alpha f(y)
\right]
}{
Z_\alpha(\theta,x)
}.
\label{eq:app-change-of-measure}
\end{equation}

By the autoregressive factorization,
\begin{align}
\log
\frac{
\pi_\theta(y\mid x)
}{
\pi_T(y\mid x)
}
&=
\log
\frac{
\prod_{t=1}^{T}\pi_\theta(y_t\mid h_t)
}{
\prod_{t=1}^{T}\pi_T(y_t\mid h_t)
}
\notag\\
&=
\sum_{t=1}^{T}
\log
\frac{
\pi_\theta(y_t\mid h_t)
}{
\pi_T(y_t\mid h_t)
}.
\label{eq:app-autoregressive-log-ratio}
\end{align}
Substituting Eq.~\eqref{eq:app-autoregressive-log-ratio} into
Eq.~\eqref{eq:app-change-of-measure} gives the exact trajectory-level
importance-weighted form of the non-constant part of
Eq.~\eqref{eq:app-projected-objective}:
\begin{equation}
\widetilde{\mathcal{J}}_{q_\theta^\star}(\theta;x)
\coloneqq
-\left( 
D_{\mathrm{KL}}
\left(
q_{\theta}^\star
\middle\|
\pi_\theta
\right)-\rho
\right)
=
-\frac{
\mathbb{E}_{y\sim\pi_\theta(\cdot\mid x)}
\left[
r_\theta(y\mid x)^\alpha
\sum_{t=1}^{T}
\log
\frac{
\pi_\theta(y_t\mid h_t)
}{
\pi_T(y_t\mid h_t)
}
\right]
}{
Z_\alpha(\theta,x)
}.
\label{eq:app-trajectory-iw-objective}
\end{equation}

Eq.~\eqref{eq:app-trajectory-iw-objective} is a sequence-level expression:
every token term in the same sampled trajectory is multiplied by the full
trajectory ratio $r_\theta(y\mid x)^\alpha$. However, OPD is optimized through
a token-local semi-gradient on student-sampled prefixes. Hence the coefficient
assigned to the token term at position $t$ should depend only on the causal
prefix $h_t=(x,y_{<t})$, rather than on the future suffix $y_{\ge t}$.

To obtain the causal token-level surrogate, decompose the trajectory ratio as
\begin{align}
r_\theta(y\mid x)^\alpha
&=
\left(
\frac{
\pi_T(y\mid x)
}{
\pi_\theta(y\mid x)
}
\right)^\alpha \notag\\
&=
\left(
\frac{
\pi_T(y_{<t}\mid x)
}{
\pi_\theta(y_{<t}\mid x)
}
\right)^\alpha
\left(
\frac{
\pi_T(y_{\ge t}\mid x,y_{<t})
}{
\pi_\theta(y_{\ge t}\mid x,y_{<t})
}
\right)^\alpha .
\end{align}
The first factor is the prefix likelihood ratio inherited from the
trajectory-level ratio in Proposition~\ref{prop:constrained-projection}. We
write
\begin{align}
r_t(y_{<t}\mid x)
&\coloneqq
r_\theta(y_{<t}\mid x)
=
\frac{
\pi_T(y_{<t}\mid x)
}{
\pi_\theta(y_{<t}\mid x)
}
\notag\\
&=
\prod_{k=1}^{t-1}
\frac{
\pi_T(y_k\mid x,y_{<k})
}{
\pi_\theta(y_k\mid x,y_{<k})
}.
\label{eq:app-prefix-ratio}
\end{align}
The empty product is $1$, so $r_1=1$. The position-wise normalizer is
\begin{equation}
Z_{\alpha,t}(\theta,x)
\coloneqq
\mathbb{E}_{y_{<t}\sim\pi_\theta(\cdot\mid x)}
\left[
r_t(y_{<t}\mid x)^\alpha
\right],
\label{eq:app-prefix-normalizer}
\end{equation}
and the normalized prefix ratio is
\begin{equation}
\tilde r_{t}(y_{<t}\mid x)
\coloneqq
\frac{
r_t(y_{<t}\mid x)^\alpha
}{
Z_{\alpha,t}(\theta,x)
}.
\label{eq:app-normalized-prefix-ratio}
\end{equation}

Replacing the full trajectory ratio in
Eq.~\eqref{eq:app-trajectory-iw-objective} with its causal prefix component
yields the token-level IW-OPD surrogate for a fixed prompt:
\begin{equation}
\mathcal{J}_{\mathrm{IW}}^\star(\theta;x)
=
\max_\theta \;
-\mathbb{E}_{y\sim\pi_\theta(\cdot\mid x)}
\left[
\sum_{t=1}^{T}
\mathrm{sg}
\left[
\tilde r_{t}(y_{<t}\mid x)
\right]
\log
\frac{
\pi_\theta(y_t\mid h_t)
}{
\pi_T(y_t\mid h_t)
}
\right].
\label{eq:app-token-iw-objective}
\end{equation}
Here $\mathrm{sg}[\cdot]$ denotes the stop-gradient operator. It makes the
normalized prefix ratio act as a detached multiplicative coefficient on the
standard token-level log-ratio term.

Averaging Eq.~\eqref{eq:app-token-iw-objective} over
$x\sim\mathcal{D}$ gives
\begin{equation}
\mathcal{J}_{\mathrm{IW}}^\star(\theta)
=
\max_\theta \;
-\mathbb{E}_{x\sim\mathcal{D},\,y\sim\pi_\theta(\cdot\mid x)}
\left[
\sum_{t=1}^{T}
\mathrm{sg}
\left[
\tilde r_{t}(y_{<t}\mid x)
\right]
\log
\frac{
\pi_\theta(y_t\mid x,y_{<t})
}{
\pi_T(y_t\mid x,y_{<t})
}
\right].
\label{eq:app-iw-opd-final}
\end{equation}
Suppressing the fixed prompt $x$ in the notation recovers
Eq.~\eqref{eq:pro_2}.
\hfill$\square$

\subsection{Standard OPD Chain Rule and Single-Step Semi-Gradient}
\label{app:standard-opd-semigradient}

We spell out the chain-rule decomposition of standard OPD and the
single-step semi-gradient used by Eq.~\eqref{eq:iw-opd-gradient}. Fix a prompt
$x$ and write $h_t=(x,y_{<t})$. We keep the conditioning on $x$ explicit in this
appendix, while the main text suppresses it when no ambiguity arises.

The per-prompt reverse-KL quantity minimized by standard OPD is
\begin{equation}
J_{\mathrm{OPD}}(\theta;x)
=
\max_\theta -D_{\mathrm{KL}}
\left(
\pi_\theta
\middle\|
\pi_T
\right)
=
-\mathbb{E}_{y\sim\pi_\theta(\cdot\mid x)}
\left[
\log
\frac{
\pi_\theta(y\mid x)
}{
\pi_T(y\mid x)
}
\right].
\end{equation}
By the autoregressive factorization,
\begin{equation}
\log
\frac{
\pi_\theta(y\mid x)
}{
\pi_T(y\mid x)
}
=
\sum_{t=1}^{T}
\log
\frac{
\pi_\theta(y_t\mid h_t)
}{
\pi_T(y_t\mid h_t)
}.
\end{equation}
Therefore,
\begin{align}
D_{\mathrm{KL}}
\left(
\pi_\theta
\middle\|
\pi_T
\right)
&=
\mathbb{E}_{y\sim\pi_\theta(\cdot\mid x)}
\left[
\sum_{t=1}^{T}
\log
\frac{
\pi_\theta(y_t\mid h_t)
}{
\pi_T(y_t\mid h_t)
}
\right]
\notag\\
&=
\sum_{t=1}^{T}
\mathbb{E}_{y_{<t}\sim\pi_\theta(\cdot\mid x)}
\left[
\mathbb{E}_{y_t\sim\pi_\theta(\cdot\mid h_t)}
\left[
\log
\frac{
\pi_\theta(y_t\mid h_t)
}{
\pi_T(y_t\mid h_t)
}
\right]
\right]
\notag\\
&=
\sum_{t=1}^{T}
\mathbb{E}_{y_{<t}\sim\pi_\theta(\cdot\mid x)}
\left[
D_{\mathrm{KL}}
\left(
\pi_\theta(\cdot\mid h_t)
\middle\|
\pi_T(\cdot\mid h_t)
\right)
\right].
\label{eq:app-standard-opd-chain-rule}
\end{align}

Next consider one local next-token KL at a fixed prefix $h_t$:
\begin{equation}
D_t(\theta;h_t)
=
D_{\mathrm{KL}}
\left(
\pi_\theta(\cdot\mid h_t)
\middle\|
\pi_T(\cdot\mid h_t)
\right).
\end{equation}
Expanding over the next-token vocabulary gives
\begin{equation}
D_t(\theta;h_t)
=
\sum_{a}
\pi_\theta(a\mid h_t)
\left[
\log \pi_\theta(a\mid h_t)
-
\log \pi_T(a\mid h_t)
\right].
\end{equation}
Differentiating this local KL while holding the sampled prefix $h_t$ fixed
yields
\begin{align}
\nabla_\theta D_t(\theta;h_t)
&=
\sum_{a}
\nabla_\theta \pi_\theta(a\mid h_t)
\left[
\log \pi_\theta(a\mid h_t)
-
\log \pi_T(a\mid h_t)
+
1
\right]
\notag\\
&=
\mathbb{E}_{a\sim\pi_\theta(\cdot\mid h_t)}
\left[
\left(
\log
\frac{
\pi_\theta(a\mid h_t)
}{
\pi_T(a\mid h_t)
}
+
1
\right)
\nabla_\theta\log\pi_\theta(a\mid h_t)
\right].
\end{align}
The $+1$ term vanishes because
\begin{equation}
\mathbb{E}_{a\sim\pi_\theta(\cdot\mid h_t)}
\left[
\nabla_\theta\log\pi_\theta(a\mid h_t)
\right]
=
\nabla_\theta
\sum_a
\pi_\theta(a\mid h_t)
=
0.
\end{equation}
Thus,
\begin{equation}
\nabla_\theta D_t(\theta;h_t)
=
\mathbb{E}_{a\sim\pi_\theta(\cdot\mid h_t)}
\left[
\log
\frac{
\pi_\theta(a\mid h_t)
}{
\pi_T(a\mid h_t)
}
\nabla_\theta\log\pi_\theta(a\mid h_t)
\right].
\label{eq:app-local-kl-gradient}
\end{equation}

Using the OPD advantage notation in Eq.~\eqref{eq:opd-advantage}, for a sampled
token $a=y_t$ we have
\begin{equation}
A_t^{\mathrm{OPD}}
:=
-
\left(
\log \pi_\theta(y_t\mid h_t)
-
\log \pi_T(y_t\mid h_t)
\right)
=
-
\log
\frac{
\pi_\theta(y_t\mid h_t)
}{
\pi_T(y_t\mid h_t)
}.
\end{equation}
Therefore, the negative local KL gradient, written in policy-gradient ascent
form, is
\begin{equation}
-\nabla_\theta D_t(\theta;h_t)
=
\mathbb{E}_{y_t\sim\pi_\theta(\cdot\mid h_t)}
\left[
A_t^{\mathrm{OPD}}
\nabla_\theta\log\pi_\theta(y_t\mid h_t)
\right].
\label{eq:app-local-opd-semigradient}
\end{equation}

The word ``semi-gradient'' is important. The chain-rule decomposition in
Eq.~\eqref{eq:app-standard-opd-chain-rule} is exact, but the token-local update
in Eq.~\eqref{eq:app-local-opd-semigradient} treats the sampled prefix
$y_{<t}$ as fixed context. A full score-function gradient of the sequence-level
reverse KL would instead couple all positions:
\begin{equation}
\nabla_\theta
D_{\mathrm{KL}}
\left(
\pi_\theta(\cdot\mid x)
\middle\|
\pi_T(\cdot\mid x)
\right)
=
\mathbb{E}_{y\sim\pi_\theta(\cdot\mid x)}
\left[
\left(
\sum_{k=1}^{T}
\log
\frac{
\pi_\theta(y_k\mid h_k)
}{
\pi_T(y_k\mid h_k)
}
\right)
\left(
\sum_{t=1}^{T}
\nabla_\theta\log\pi_\theta(y_t\mid h_t)
\right)
\right].
\label{eq:app-full-opd-gradient}
\end{equation}
Standard OPD practice instead uses the single-step token-local update direction.
With the sign convention of Eq.~\eqref{eq:opd-advantage}, this gives
\begin{equation}
\nabla_\theta J_{\mathrm{OPD}}(\theta)
\approx
\mathbb{E}_{x\sim\mathcal{D},\,y\sim\pi_\theta(\cdot\mid x)}
\left[
\sum_{t=1}^{T}
A_t^{\mathrm{OPD}}
\nabla_\theta\log\pi_\theta(y_t\mid h_t)
\right],
\label{eq:app-standard-opd-semigradient}
\end{equation}
which is the standard OPD form in the main text.

We now apply the same single-step rule to the ideal IW-OPD surrogate in
Proposition~\ref{prop:iw-opd-objective}. For a fixed prompt, Eq.~\eqref{eq:pro_2}
can be written as
\begin{equation}
J_{\mathrm{IW}}^\star(\theta;x)
=
\max_\theta \;
-\mathbb{E}_{y\sim\pi_\theta(\cdot\mid x)}
\left[
\sum_{t=1}^{T}
\mathrm{sg}
\left[
\tilde r_{t}(y_{<t}\mid x)
\right]
\log
\frac{
\pi_\theta(y_t\mid h_t)
}{
\pi_T(y_t\mid h_t)
}
\right],
\label{eq:app-iw-opd-objective-for-gradient}
\end{equation}
where the notation follows the main text:
\begin{equation}
r_t
:=
r_\theta(y_{<t}\mid x)
=
\frac{
\pi_T(y_{<t}\mid x)
}{
\pi_\theta(y_{<t}\mid x)
},
\qquad
Z_{\alpha,t}(\theta,x)
=
\mathbb{E}_{y_{<t}\sim\pi_\theta(\cdot\mid x)}
\left[
r_t
\right],
\qquad
\tilde r_{t}
=
\frac{
r_t
}{
Z_{\alpha,t}
}.
\end{equation}

When differentiating the local next-token term at position $t$, all
prefix-determined quantities are treated as fixed context. This includes the
sampled prefix $y_{<t}$, the prefix ratio $r_t$, the normalizer
$Z_{\alpha,t}$, and the normalized prefix ratio $\tilde r_{t}$. The
stop-gradient operator in Eq.~\eqref{eq:app-iw-opd-objective-for-gradient}
enforces exactly this convention.

Detaching the prefix ratio is necessary for preserving the single-step OPD
semi-gradient. Indeed, for fixed $\alpha$,
\begin{align}
\log r_t
&=
\alpha
\sum_{k<t}
\left(
\log \pi_T(y_k\mid h_k)
-
\log \pi_\theta(y_k\mid h_k)
\right)
\notag\\
&=
\alpha
\sum_{k<t}
A_k^{\mathrm{OPD}}.
\label{eq:app-prefix-ratio-log}
\end{align}
Since the teacher is fixed, differentiating through the ratio would introduce
prefix score terms:
\begin{equation}
\nabla_\theta \log r_t
=
-\alpha
\sum_{k<t}
\nabla_\theta
\log\pi_\theta(y_k\mid h_k).
\end{equation}
Such terms couple the token-$t$ update to earlier sampled actions and recover a
sequence-level credit-assignment estimator rather than the token-local OPD
semi-gradient. Differentiating $Z_{\alpha,t}$ would similarly require gradients
through both the prefix sampling distribution and the prefix likelihood ratio.
Therefore, both $r_t$ and $Z_{\alpha,t}$ are detached in the
single-step update.

With these detached prefix quantities, the local IW-OPD update is simply the
standard OPD update multiplied by the normalized prefix ratio. Thus,
\begin{align}
\nabla_\theta J_{\mathrm{IW}}^\star(\theta)
&\approx
\mathbb{E}_{x\sim\mathcal{D},\,y\sim\pi_\theta(\cdot\mid x)}
\left[
\sum_{t=1}^{T}
A_t^{\mathrm{IW\mbox{-}OPD}}
\nabla_\theta\log\pi_\theta(y_t\mid h_t)
\right],
\label{eq:app-iw-opd-semigradient}
\end{align}
where
\begin{align}
A_t^{\mathrm{IW\mbox{-}OPD}}
&=
\mathrm{sg}
\left[
\tilde r_{t}
\right]
A_t^{\mathrm{OPD}}
\notag\\
&=
-
\mathrm{sg}
\left[
\tilde r_{t}
\right]
\left(
\log \pi_\theta(y_t\mid h_t)
-
\log \pi_T(y_t\mid h_t)
\right).
\label{eq:app-iw-opd-advantage}
\end{align}
This is the semi-gradient form of Eq.~\eqref{eq:iw-opd-gradient}, with the
stop-gradient convention inherited from Eq.~\eqref{eq:pro_2}.
\hfill$\square$

\section{Algorithm}
\label{sec:iwopd-algorithm}

For clipped PPO, let $\pi_0$ be the frozen rollout policy for the current batch, and use
$\eta_t(\theta)=\pi_\theta(y_t\mid x,y_{<t})/\pi_0(y_t\mid x,y_{<t})$
for the PPO ratio. The inner update minimizes
\begin{equation}
\mathcal{L}_{\mathrm{IW\text{-}OPD}}(\theta)
=
-
\mathbb{E}_{x,y,t}
\left[
\min\left(
\eta_t(\theta)A_t^{\mathrm{IW\text{-}OPD}},
\operatorname{clip}\!\left(
\eta_t(\theta),
1-\epsilon_{\mathrm{clip}},
1+\epsilon_{\mathrm{clip}}
\right)
A_t^{\mathrm{IW\text{-}OPD}}
\right)
\right],
\label{eq:iwopd-loss}
\end{equation}
where the expectation is over valid response tokens.

\begin{algorithm}[H]
\caption{IW-OPD: Importance-Weighted On-Policy Distillation}
\label{alg:iw-opd}
\setstretch{1.3}
\begin{algorithmic}[1]
\REQUIRE Student $\pi_\theta$, teacher $\pi_T$, prompt distribution $\mathcal{D}$, amplification $\gamma$ (default $0.5$), PPO clip $\epsilon_{\mathrm{clip}}$, stabilizer $\varepsilon$.
\STATE Initialize rollout policy $\pi_0 \leftarrow \pi_\theta$.
\FOR{each training iteration}
    \STATE Sample $x \sim \mathcal{D}$ and generate $y=(y_1,\ldots,y_T)$ from $\pi_0(\cdot\mid x)$.
    \STATE Cache $\ell_{0,t}\leftarrow \log \pi_0(y_t\mid x,y_{<t})$ and $\ell_{T,t}\leftarrow \log \pi_T(y_t\mid x,y_{<t})$ for all valid tokens $t$.
    \STATE Set $A_t^{\mathrm{OPD}}\leftarrow \mathrm{sg}[\ell_{T,t}-\ell_{0,t}]$ for all valid tokens $t$.
    \STATE Set $\tilde{r}^\mathrm{IW-OPD}_t\leftarrow 1+\gamma\left(1-
    \frac{\sum_{k<t}\left|A_k^{\mathrm{OPD}}\right|}
    {\sum_{k<T}\left|A_k^{\mathrm{OPD}}\right|}\right)$ for all valid tokens $t$.
    \STATE Set $A_t^{\mathrm{IW\text{-}OPD}}\leftarrow \mathrm{sg}[\tilde{r}^\mathrm{IW-OPD}_t]A_t^{\mathrm{OPD}}$ for all valid tokens $t$.
    \FOR{several PPO inner steps}
        \STATE Update $\theta$ by minimizing Eq.~\eqref{eq:iwopd-loss}, using $\eta_t(\theta)=\exp(\log \pi_\theta(y_t\mid x,y_{<t})-\ell_{0,t})$.
    \ENDFOR
    \STATE Refresh rollout policy: $\pi_0 \leftarrow \pi_\theta$.
\ENDFOR
\end{algorithmic}
\end{algorithm}

\section{Experimental Setup and Hyperparameters}
\label{app:experimental-details}

This section reports the implementation details used for the experiments in \S\ref{sec:experiments}. All OPD variants are implemented in the same \texttt{verl}-based PPO training pipeline. The student samples responses on-policy; student and teacher log probabilities are then evaluated on the sampled response tokens. No learned reward model is used: the token-level OPD or IW-OPD advantages are passed directly into the clipped PPO surrogate.

\subsection{Models and Data}

\begin{table}[ht]
\centering
\small
\caption{Model and data configurations used in the main experiments.}
\label{tab:app-model-data}
\begin{tabular}{p{0.25\linewidth}p{0.68\linewidth}}
\toprule
\textbf{Component} & \textbf{Configuration} \\
\midrule
Students & Qwen3-4B, Qwen3-1.7B, and Qwen3-0.6B. \\
Teachers & Qwen3-4B-Instruct-2507 for the large-overlap setting; Qwen3-30B-A3B-Instruct-2507 for the small-overlap setting. \\
Math training data & DeepMath problems filtered to difficulty level $\geq 6$ (approximately 57K prompts). \\
Code training data & Eurus-RL-Code (approximately 25K prompts). \\
Validation during training & AIME 2024 and AIME 2025, evaluated every 10 training steps. \\
Final evaluation & AIME 2024, AIME 2025, and HMMT 2025 for math; HumanEval+ and MBPP+ for code. \\
\bottomrule
\end{tabular}
\end{table}

For all Qwen3 models, we use the chat template with thinking disabled. Prompts longer than the context budget are filtered rather than truncated, and response tokens beyond the generated answer mask are excluded from all OPD and IW-OPD computations.

\subsection{Training Hyperparameters}
\label{app:training-parameters}

\begin{table}[ht]
\centering
\small
\caption{Default training hyperparameters. Unless noted otherwise, the same values are used for OPD and IW-OPD within each model--teacher setting.}
\label{tab:app-training-hparams}
\begin{tabular}{p{0.35\linewidth}p{0.55\linewidth}}
\toprule
\textbf{Hyperparameter} & \textbf{Value} \\
\midrule
Training framework & \texttt{verl} PPO trainer with vLLM rollouts \\
Nodes / GPUs & 4 nodes, 32 GPUs \\
Learning rate & $1\times 10^{-5}$ \\
Training batch size & 1024 prompts \\
PPO mini-batch size & 1024 \\
PPO micro-batch size & 1 per GPU \\
PPO epochs per rollout batch & 1 \\
PPO clipping range & 0.2 \\
Dual-clip constant & 3.0 \\
Loss aggregation & Token mean \\
Maximum prompt length & 2048 tokens \\
Maximum response length & 16384 tokens \\
Rollout samples per prompt & 1 during training \\
Training decoding & Temperature $1.0$, top-$p$ $1.0$ \\
Validation decoding & Temperature $1.0$, top-$p$ $1.0$, 32 samples per prompt for math validation \\
Optimizer warmup & 0 warmup ratio \\
Entropy coefficient & 0 \\
KL reward penalty & Disabled \\
Auxiliary KL loss & Low-variance KL form, coefficient 0 \\
Rollout importance correction & Token-level correction with threshold 5.0 \\
Checkpoint / evaluation frequency & Save every 10 steps; validate every 10 steps \\
\bottomrule
\end{tabular}
\end{table}

Epoch budgets follow the corresponding model-scale scripts and are held fixed across OPD and IW-OPD within each setting. All reported comparisons use the same data order and seed set across methods.

\subsection{Method-Specific Parameters}

Finally, IW-OPD perform an interpolation with the original OPD to balance these two effects:
\begin{equation}
A^\mathrm{IW-OPD}_t =
\mathrm{sg}[\tilde{r}^\mathrm{IW-OPD}_t] \cdot A^\mathrm{OPD}_t =
\left( 1+\gamma\left(1-
\frac{\sum_{k<t}\left|A_k^{\mathrm{OPD}}\right|}
{\sum_{k<T}\left|A_k^{\mathrm{OPD}}\right|}\right) \right) \cdot A^\mathrm{OPD}_t,
\end{equation}
where higher $\gamma\geq0$ indicates a higher contribution from the teacher-informed importance weights. In practice, we find $\gamma=0.5$ is a good default choice.

\subsection{Evaluation Protocol}

For math benchmarks, we generate 32 responses per problem with vLLM using temperature $1.0$, top-$p$ $1.0$, maximum generation length 16384, and seed-matched sampling across methods. Each prompt appends the instruction: ``Please reason step by step, and put your final answer within \textbackslash boxed\{\}.'' We extract the final boxed answer and evaluate it with symbolic equivalence checking. Tables report the aggregation specified in their captions, e.g., best@32 or mean@32.

For code benchmarks, we use the EvalPlus evaluation suite for HumanEval+ and MBPP+. Greedy single-sample evaluation is used for the reported pass-rate results.

Checkpoints with subscript $_{10}$ in the main tables are evaluated at training step 10. Converged OPD and IW-OPD checkpoints are selected using the same validation protocol within each model--teacher setting and are then evaluated on the held-out math and code benchmarks.

\section{Combination experiments with other reward design methods}
\label{app:exopd}

\begin{table}[H]
\centering
\caption{Combination results with ExOPD; both methods are distilled from Qwen3-30B-A3B. Math results are reported as mean@32 accuracy (\%). IW-ExOPD denotes the combination of IW-OPD and ExOPD.
\textbf{Bold} indicates the best result within each student group.}
\label{tab:exopd_results}
\resizebox{\textwidth}{!}{
\begin{tabular}{ll|ccc|cc|A}
\toprule
\multirow{2}{*}{\textbf{Student}} & \multirow{2}{*}{\textbf{Method}} & \multicolumn{3}{c|}{\textbf{Math}} & \multicolumn{2}{c|}{\textbf{Code}} & \multirow{2}{*}{\textbf{Avg}} \\
& & AIME24 & AIME25 & HMMT25 & \phantom{0}HE+\phantom{0} & MBPP+ & \\
\midrule
\multicolumn{2}{l|}{\textcolor{gray}{\textit{Teacher Model}}}
& \textcolor{gray}{74.7}
& \textcolor{gray}{62.8}
& \textcolor{gray}{44.2}
& \phantom{0}\textcolor{gray}{86.6}\phantom{00}
& \textcolor{gray}{75.1}
& \textcolor{gray}{68.7}  \\
\midrule
\multirow{4}{*}{Qwen3-4B}
& Base
& 23.1
& 21.4
& 10.0
& \phantom{0}75.3\phantom{00}
& 64.5
& 38.9 \\

& OPD
& 55.3
& 48.0
& 27.1
& \phantom{0}77.2\phantom{00}
& 69.1
& 55.3 \\

& ExOPD
& 57.9
& 50.1
& 31.7
& \phantom{0}78.9\phantom{00}
& 70.2
& 57.8 \\

& \cellcolor{orgLight}IW-ExOPD
& \cellcolor{orgLight}\cimp{\textbf{59.4}}{+1.5}
& \cellcolor{orgLight}\cimp{\textbf{51.7}}{+1.6}
& \cellcolor{orgLight}\cimp{\textbf{32.0}}{+0.3}
& \cellcolor{orgLight}\phantom{0}\cimp{\textbf{80.1}}{+1.1}\phantom{00}
& \cellcolor{orgLight}\cimp{\textbf{71.0}}{+0.8}
& \cellcolor{orgLight}\cimp{\textbf{58.8}}{+1.0} \\
\midrule
\multirow{4}{*}{Qwen3-1.7B}
& Base
& 13.4
& 11.0
& 6.8
& \phantom{0}59.6\phantom{00}
& 52.5
& 28.7 \\

& OPD
& 34.6
& 28.7
& 15.5
& \phantom{0}64.6\phantom{00}
& 53.7
& 39.4 \\

& ExOPD
& 37.6
& 31.8
& 16.8
& \phantom{0}67.2\phantom{00}
& 55.0
& 41.7 \\

& \cellcolor{orgLight}IW-ExOPD
& \cellcolor{orgLight}\cimp{\textbf{38.9}}{+1.3}
& \cellcolor{orgLight}\cimp{\textbf{33.2}}{+1.4}
& \cellcolor{orgLight}\cimp{\textbf{18.3}}{+1.5}
& \cellcolor{orgLight}\phantom{0}\cimp{\textbf{68.5}}{+1.3}\phantom{00}
& \cellcolor{orgLight}\cimp{\textbf{57.4}}{+2.4}
& \cellcolor{orgLight}\cimp{\textbf{43.2}}{+1.5} \\
\bottomrule
\end{tabular}
}
\vspace{-5pt}
\end{table}

\paragraph{IW-OPD is orthogonal to reward design.} IW-OPD changes how an already-computed OPD signal is allocated across positions, so it is not tied to a particular advantage or reward design. As one example, ExOPD~\citep{yang2026gopd} reformulates OPD as a reinforcement-learning problem with a KL constraint, separates out the reward term, and improves exploration by scaling that reward with a fixed hyperparameter $\lambda$. Since IW-OPD supplies prefix-level importance weights, we can apply the same idea on top of ExOPD by making the reward scale prefix-dependent. We call this combination IW-ExOPD. As shown in Table~\ref{tab:exopd_results}, IW-ExOPD improves over ExOPD for both Qwen3-4B and Qwen3-1.7B students. This suggests that prefix-level importance weighting is orthogonal to reward design, rather than being tied only to vanilla OPD.

\section{Limitations}
At convergence, $\tilde{r}_t$ approaches $1-t/T$ rather than becoming perfectly uniform---a mild residual non-uniformity. Cumulative prefix discrepancy is a conservative prefix-compatibility proxy induced by the prefix likelihood-ratio principle, not an exact density-ratio correction. Experiments are conducted at the 4B student scale; validation at larger scales remains future work.

\end{document}